\documentclass[letterpaper,10pt,journal,twoside]{IEEEtran}

\usepackage{graphicx}
\usepackage{adjustbox}
\usepackage{amsmath}
\usepackage{amssymb}
\usepackage{multirow}
\usepackage{float}
\usepackage{booktabs}
\usepackage[nocompress]{cite}
\usepackage{color}
\usepackage[ruled,vlined,linesnumbered]{algorithm2e}
\usepackage{algpseudocode}
\usepackage{placeins}
\usepackage{array}
\usepackage{listings}
\usepackage{tikz}
\usetikzlibrary{shapes.geometric, arrows.meta, calc, angles, quotes, decorations.pathreplacing}
\usepackage{pgfplots}
\usepgfplotslibrary{groupplots}
\pgfplotsset{compat=1.18}

\definecolor{COLORSPARSITY}{HTML}{2E86AB}
\definecolor{COLORDECISIVE}{HTML}{D62828}
\definecolor{COLORARROW}{HTML}{333333}
\usepackage{makecell}
\usepackage[hidelinks]{hyperref}

\title{
DSGN: A Dual-Stream Gated Network for Scalable Simultaneous Surveillance and Containment
}

\author{Heng Zhang$^{1,\dagger}$, Ziyang Zuo$^{1,\dagger}$, Guoxiang Zhao$^{2,*}$, and Xiaoqiang Ren$^{1,2,3}$%
    \thanks{The work was supported in part by the National Natural Science Foundation of China under Grant 62336005, 62403300, 62421004 and 62461160313,  and the Shanghai Municipal Commission of Education under Grant 24SG38.}%
    \thanks{$^{\dagger}$Heng Zhang and Ziyang Zuo contributed equally to this work.}%
    \thanks{$^{1}$Heng Zhang, Ziyang Zuo, and Xiaoqiang Ren are with the School of Mechatronic Engineering and Automation, Shanghai University, Shanghai 200444, China (e-mail: {\tt\small heng-zhang@shu.edu.cn; zzysama@shu.edu.cn; xqren@shu.edu.cn}).}%
    \thanks{$^{2}$Guoxiang Zhao and Xiaoqiang Ren are with the School of Future Technology, Shanghai University, Shanghai 200444, China (e-mail: {\tt\small gxzhao@shu.edu.cn}).}%
    \thanks{$^{3}$Xiaoqiang Ren is with the National Key Laboratory of Maritime Unmanned Swarm Intelligence, Shanghai University, Shanghai 200444, China.}%
    \thanks{$^{*}$Corresponding author: Guoxiang Zhao.}
}

\begin{document}

\bstctlcite{RALReferenceControl}

\maketitle

    \begin{abstract}

        The simultaneous surveillance and containment (SSC) problem arises in multi-robot applications such as perimeter security and environmental monitoring, where robots must survey a region of interest and coordinate to enclose targets once they are detected.
Surveillance favors spatial dispersion for coverage, whereas containment requires target-centric cohesion for enclosure.
This conflict makes heuristic switching unstable, and the instability becomes more pronounced as larger robotic networks introduce more inter-agent and agent-target relations.
End-to-end DRL reduces reliance on hand-crafted switching, but conventional single-stream policies often mix surveillance and containment cues in one shared feature space that could entangle dispersion-oriented and target-centric signals and weaken task transitions.
In this paper, we propose the Dual-Stream Gated Network (DSGN), an architecture designed to address scale-amplified task-transition difficulties by separating surveillance and containment into specialized representation streams and coordinating them through a context-aware gating mechanism.
A probabilistic prior further regularizes the gating process, reducing ambiguous intermediate decisions and supporting clearer behavioral transitions under large-scale interaction complexity.
Evaluations show that DSGN maintains strong task efficacy and scalability under constant-density cross-scale settings.
Additional ablation and phase-specific evaluations show that the proposed dynamic gating mechanism and probabilistic prior support effective task arbitration and reliable mode switching in both pure surveillance and task-switching scenarios.
The source code is available at 
\url{https://github.com/zuozysama/DSGN}.

    \end{abstract}


    \section{INTRODUCTION}
    \IEEEPARstart{A}{s} robotic autonomy matures, large-scale multi-robot systems are increasingly deployed across diverse domains, ranging from environmental monitoring and industrial automation to critical infrastructure maintenance~\cite{yang2018grand, dorigo2021swarm}.
A representative example is autonomous fishing fleets~\cite{vanhee2018autonomous}, where robots must survey a region of interest and coordinate their response to mobile, non-cooperative entities.
These scenarios necessitate two fundamental capabilities: persistent area surveillance to locate multiple targets~\cite{liu2021cooperative, wang2023spatio}, and coordinated containment operations to physically enclose them~\cite{zhang2025terl}.
\textcolor{black}{Although searching and containment have each been extensively studied, they are generally formulated separately.
The simultaneous surveillance and containment (SSC) problem considered here couples them by requiring a multi-robot system to balance continuous area coverage with timely containment of targets.}

Centralized coordination~\cite{qu2020scalable} can exploit global information, but its dependence on the global system state makes it prone to the curse of dimensionality and limits its applicability to large-scale robotic networks.
Distributed methods~\cite{busoniu2008comprehensive, weintraub2020introduction} mitigate this limitation by allowing robots to coordinate through local observations and inter-robot interactions.
\textcolor{black}{Existing distributed methods relevant to the containment component of SSC have primarily focused on target interception and containment.}
Differential game methods~\cite{weintraub2020introduction} model interactions between pursuers and evaders as coupled optimization problems, but they can be difficult to apply when environment dynamics are hard to model.
Bio-inspired heuristic methods~\cite{wang2022pursuit} use manually designed interaction rules, which are intuitive to implement but can be hard to scale across varying team sizes, environments, and target behaviors.
Deep reinforcement learning (DRL) methods~\cite{zhang2022multi} have therefore been introduced to learn containment policies directly from interactions, reducing dependence on explicit dynamics modeling and manually specified interaction rules.
These studies cover both single-target containment~\cite{wang2020cooperative} and multi-target settings~\cite{yang2023large}, and recent work~\cite{zhang2025terl} has scaled up multi-target coordination to dozens of robots by leveraging attention-based architectures.
\textcolor{black}{The minimal capture time in the graph-based pursuit-evasion formulation has been investigated in~\cite{bonato2009capture}.}
However, in practical multi-robot applications such as environmental monitoring~\cite{wang2023spatio} and autonomous fishing~\cite{vanhee2018autonomous}, targets may not be immediately known in the region of interest.
This means that containment alone is insufficient and that persistent surveillance is also required to locate targets before coordinated interception or enclosure~\cite{liu2021cooperative, wang2023spatio}.

Multi-robot surveillance has been widely studied for persistent coverage and target localization.
\textcolor{black}{Classical visibility-based pursuit-evasion studies established complete search strategies and searchability conditions in polygonal and curved environments~\cite{guibas1999visibility,park2001visibility,lavalle2001curved}.
Multi-robot target search has also addressed scalable planning for moving targets~\cite{hollinger2009efficient}, limited sensing and environment geometry~\cite{agrawal2025searching}, and DRL-based mobile-target search~\cite{ye2025deep}.
These studies primarily address target search and detection, whereas SSC couples persistent surveillance with post-detection geometric containment.}
A straightforward way to combine the two behaviors is to formulate the transition as a switched or hybrid control problem~\cite{liberzon2003switching,branicky1998stability}.
However, specifying analytical switching conditions for coupled target search and interception in SSC is difficult because the desired behavior depends jointly on coverage status, target observations, robot distribution, and containment geometry.

\textcolor{black}{An end-to-end DRL paradigm offers an alternative to discrete logic for the SSC problem by learning decentralized policies directly from robot local observations~\cite{mnih2015human, zhang2022multi}.}
However, conventional unified architectures often encode surveillance and containment cues in a shared feature space, where dispersion-oriented coverage signals and target-centric enclosure signals can introduce conflicting optimization demands~\cite{yu2020gradient}.
Within this end-to-end paradigm, a commonly used design is to organize a single policy with specialized representation streams and coordinate them through a gating mechanism, as explored in mixture-of-experts~\cite{jacobs1991adaptive} and modular reinforcement learning~\cite{andreas2017modular}.
Fixed or hand-crafted gating mechanisms can be difficult to adapt as observations and task demands change, while naive adaptive gating may incur chattering that could cause task failures.

To address this challenge, we propose the Dual-Stream Gated Network (DSGN).
DSGN separates state processing into two specialized streams.
A \textit{spatial-explorative stream} reasons about coverage gaps for surveillance, and a \textit{tactical-exploitative stream} computes geometric formations for containment.
A context-aware gating mechanism coordinates the two streams to produce a unified policy for surveillance and containment.
We further regularize this gating process with a Beta-distribution prior to stabilize switching decisions.
Numerical experiments show that DSGN improves the balance between surveillance and containment and maintains performance as the number of agents increases.

The key contributions of this work are as follows:
1) \textbf{Task-Disentangled Representation for SSC:}
    Built on transformer-based relational encoding, DSGN separates surveillance and containment into two task-specific attention streams, which reduce interference between dispersion-oriented coverage and target-centered containment within a unified end-to-end policy.
2) \textbf{Robust Distributed Task Arbitration:}
    The context-aware gating mechanism and Beta-distribution regularization enable robust switching between surveillance and containment.
    Together, these components support stable and decisive task transitions when target observations are intermittent or task demands change.
3) \textbf{Scalable Multi-Robot Coordination Performance:}
Across constant-density cross-scale evaluations, DSGN maintains high task efficacy, execution optimality, safe coordination, and stable formation quality as team and environment scales increase.
Additional phase-specific evaluations further demonstrate its task-switching and surveillance capabilities.

    \section{Problem Formulation}\label{sec:problem-formulation}
    We consider a multi-robot system operating in a bounded continuous domain \(\mathcal{W}\subset\mathbb{R}^2\).
Let \([N]=\{1,\dots,N\}\), \([M]=\{1,\dots,M\}\), \([L]=\{1,\dots,L\}\), and \([K]=\{1,\dots,K\}\) denote the index sets of pursuers, evaders, obstacles, and surveillance nodes, respectively.
At time \(t\), the pursuer and evader position sets are \(\mathcal{P}^t=\{p_n^t\in\mathcal{W}\mid n\in[N]\}\) and \(\mathcal{E}^t=\{e_m^t\in\mathcal{W}\mid m\in[M]\}\).
Each static circular obstacle is represented by \((b_\ell,r_\ell)\), where \(b_\ell\in\mathcal{W}\), \(r_\ell>0\), and \(\ell\in[L]\), giving \(\mathcal{B}=\{(b_\ell,r_\ell)\mid \ell\in[L]\}\).
The fixed surveillance-node set is \(\mathcal{C}=\{c_k\in\mathcal{W}\mid k\in[K]\}\), which provides a reference discretization of \(\mathcal{W}\) for measuring persistent surveillance.
Node \(c_k\in\mathcal{C}\) is considered visited at time \(t\) if \(\|p_n^t-c_k\|\le R_{\text{visit}}\) for some pursuer \(n\in[N]\), where \(R_{\text{visit}}>0\) is the node-visit threshold.
The idleness \(d_k^t\ge0\) of node \(c_k\) records the elapsed time since its most recent visit.
The mission requires pursuers to maintain persistent surveillance over \(\mathcal{W}\) while containing all evaders.

\subsection{Dec-POMDP Formulation and System Abstraction}

We formulate the multi-robot coordination problem as a decentralized partially observable Markov decision process (Dec-POMDP)~\cite{oliehoek2016concise}, defined by \(\langle [N],\mathcal{S},\mathcal{A},\mathcal{T},\mathcal{R},\Omega,\mathcal{O},\gamma\rangle\).
Here, \([N]\) is the set of pursuers, \(\mathcal{S}\) is the global state space, \(\mathcal{A}\) is the discrete action space, \(\mathcal{T}\) is the transition function, \(\mathcal{R}\) is the team reward function, \(\Omega\) is the observation space, \(\mathcal{O}\) is the observation function, and \(\gamma\in[0,1)\) is the discount factor.
At time \(t\), the global state \(s^t\in\mathcal{S}\) consists of pursuer states \(\{x_n^t\}_{n\in[N]}\), evader states \(\{y_m^t\}_{m\in[M]}\), static obstacles \(\mathcal{B}\), and node idleness values \(\{d_k^t\}_{k\in[K]}\).
The pursuer state is \(x_n^t=(p_n^t,v_n^t,\psi_n^t)\), where \(v_n^t\) and \(\psi_n^t\) denote the linear speed and heading of pursuer \(n\).
The evader state is \(y_m^t=(e_m^t,u_m^t,\phi_m^t)\), where \(u_m^t\) and \(\phi_m^t\) denote the linear speed and heading of evader \(m\).
Each pursuer selects a discrete kinematic command \(a_n^t=(a_{\text{acc},n}^t,\omega_n^t)\in\mathcal{A}\), where \(a_{\text{acc},n}^t\) and \(\omega_n^t\) denote the linear acceleration and angular velocity commands, respectively.
With time step \(\Delta t\), pursuer \(n\) follows the kinematic update
\(p_n^{t+1}=p_n^t+v_n^t\Delta t[\cos\psi_n^t,\sin\psi_n^t]^\top\),
\(v_n^{t+1}=v_n^t+a_{\text{acc},n}^t\Delta t\), and
\(\psi_n^{t+1}=\psi_n^t+\omega_n^t\Delta t\).
Due to decentralized execution, pursuer \(n\in[N]\) receives a local observation \(o_n^t=\mathcal{O}(s^t,n)\in\Omega\), including its own state and nearby teammates, evaders, and obstacles within the sensing radius \(R_{\text{sense}}>0\).
The locally visible evader set is \(\mathcal{E}_{\text{vis}}^{n,t}=\{m\in[M]\mid \|p_n^t-e_m^t\|\le R_{\text{sense}}\}\).
When evaders are detected by any pursuer, their detections are shared within the team to form the aggregated observed evader set \(\mathcal{E}_{\text{obs}}^{t}=\bigcup_{n\in[N]}\mathcal{E}_{\text{vis}}^{n,t}\).
Relative position, distance, bearing, and relative velocity are treated as deterministic quantities derived from these observations.

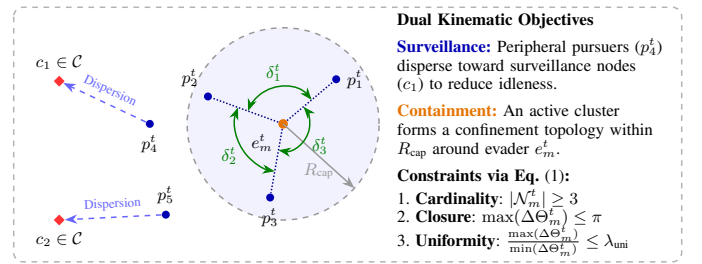
\begin{figure}[H]
    \centering

    \resizebox{\columnwidth}{!}{%
        \begin{tikzpicture}[
            >=Stealth,
            pursuer/.style={circle, fill=blue!70!black, inner sep=1.5pt},
            evader/.style={circle, fill=orange!90!black, inner sep=1.8pt},
            surveil/.style={diamond, fill=red!80, inner sep=1.5pt},
            font=\small
        ]

            \coordinate (Center) at (0,0);

            \draw[dashed, thick, gray!80, fill=blue!5] (Center) circle (1.8cm);
            \node[evader, label=below left:{$e_m^t$}] (E) at (Center) {};
            \draw[->, thick, gray!80] (Center) -- (-42:1.8cm)
            node[midway, below] {$R_{\text{cap}}$};

            \node[pursuer, label=right:{$p_1^t$}] (P1) at (40:1.3) {};
            \node[pursuer, label=above left:{$p_2^t$}] (P2) at (160:1.5) {};
            \node[pursuer, label=below:{$p_3^t$}] (P3) at (260:1.4) {};

            \draw[densely dotted, thick, blue!50!black] (E) -- (P1);
            \draw[densely dotted, thick, blue!50!black] (E) -- (P2);
            \draw[densely dotted, thick, blue!50!black] (E) -- (P3);

            \pic [draw, <->, "$\delta_1^t$", angle radius=0.7cm,
            angle eccentricity=1.4, green!50!black, thick]
                {angle = P1--E--P2};
            \pic [draw, <->, "$\delta_2^t$", angle radius=0.9cm,
            angle eccentricity=1.3, green!50!black, thick]
                {angle = P2--E--P3};
            \pic [draw, <->, "$\delta_3^t$", angle radius=0.5cm,
            angle eccentricity=1.6, green!50!black, thick]
                {angle = P3--E--P1};

            \node[surveil, label=above:{$c_1 \in \mathcal{C}$}] (C1) at (-4.2, 0.8) {};
            \node[pursuer, label=below:{$p_4^t$}] (P4) at (-2.5, 0.0) {};
            \draw[->, dashed, thick, blue!60] (P4) -- (C1)
            node[midway, sloped, above, font=\scriptsize] {Dispersion};

            \node[surveil, label=below:{$c_2 \in \mathcal{C}$}] (C2) at (-4.2, -1.8) {};
            \node[pursuer, label=above:{$p_5^t$}] (P5) at (-2.2, -1.7) {};
            \draw[->, dashed, thick, blue!60] (P5) -- (C2)
            node[midway, sloped, above, font=\scriptsize] {Dispersion};

            \node[anchor=west, text width=5.0cm, align=left, inner sep=1pt] at (2.1, -0.2) {
                \textbf{Dual Kinematic Objectives}\\[1.5mm]
                \textcolor{blue!70!black}{\textbf{Surveillance:}} Peripheral pursuers ($p_4^t$) disperse toward surveillance nodes ($c_1$) to reduce idleness.\\[1.5mm]
                \textcolor{orange!90!black}{\textbf{Containment:}} An active cluster forms a confinement topology within $R_{\text{cap}}$ around evader $e_m^t$.\\[1.5mm]
                \textbf{Constraints via Eq.~\eqref{eq:capture_cond}:}\\[0.8mm]
                1. \textbf{Cardinality}: $|\mathcal{N}_m^t| \ge 3$\\
                2. \textbf{Closure}: $\max(\Delta \Theta_m^t) \le \pi$\\
                3. \textbf{Uniformity}: $\frac{\max(\Delta \Theta_m^t)}{\min(\Delta \Theta_m^t)} \le \lambda_{\text{uni}}$
            };

            \draw[dashed, thick, gray!60, rounded corners=5pt]
            ([xshift=-3mm, yshift=0.8mm]current bounding box.north west)
            rectangle
            ([xshift=3mm, yshift=-0.8mm]current bounding box.south east);

        \end{tikzpicture}%
    }
    \vspace{-6mm}    
    \caption{Visualizing the dual kinematic objectives.}
    \label{fig:kinematic_objectives}
    \vspace{-6mm}    
\end{figure}

\subsection{Kinematic Objectives of the SSC Problem}

The SSC problem couples two kinematically different objectives, as illustrated in Fig.~\ref{fig:kinematic_objectives}.
The surveillance objective requires pursuers to disperse over \(\mathcal{W}\) to reduce node idleness, whereas the containment objective requires pursuers to contract around evaders and form target-centered confinement.
For evader \(m\in[M]\), the active containment cluster at time \(t\) is \(\mathcal{N}_m^t=\{n\in[N]\mid \|p_n^t-e_m^t\|\le R_{\text{cap}}\}\), where \(R_{\text{cap}}>0\) is the capture radius.
For each pursuer \(n\in\mathcal{N}_m^t\), let \(\theta_n^t\) denote its bearing angle with respect to evader \(m\).
After sorting these bearings in ascending order, let \(\Delta\Theta_m^t=\{\delta_1^t,\dots,\delta_{|\mathcal{N}_m^t|}^t\}\) be the set of adjacent angular gaps, where \(\delta_j^t=\theta_{j+1}^t-\theta_j^t\) for \(j=1,\dots,|\mathcal{N}_m^t|\), and \(\theta_{|\mathcal{N}_m^t|+1}^t=\theta_1^t+2\pi\).
Evader \(m\) is contained at time \(t\) if the active cluster satisfies:
\begin{equation}
    \label{eq:capture_cond}
    \begin{split}
        \mathbb{I}_{\text{cap}}(e_m^t) \iff
        &\left(|\mathcal{N}_m^t|\ge 3\right)
        \land
        \left(\max(\Delta\Theta_m^t)\le \pi\right) \\
        &
        \land
        \left(
        \frac{\max(\Delta\Theta_m^t)}
             {\min(\Delta\Theta_m^t)}
        \le \lambda_{\text{uni}}
        \right),
    \end{split}
\end{equation}
where \(\lambda_{\text{uni}}\ge1\) is a uniformity coefficient that enforces a balanced angular distribution, with \(\lambda_{\text{uni}}=1\) corresponding to equal angular dispersion among pursuers.
This condition ensures both closure of the evader escape cone and stability of the multi-robot containment formation.
The objective of SSC is therefore to maintain low surveillance idleness while satisfying the containment condition in Eq.~\eqref{eq:capture_cond} for all evaders.
Since this task objective cannot be directly optimized by reinforcement learning, the learned policy is trained through the expected discounted return \(J(\pi)=\mathbb{E}_{\pi}[\sum_{t=0}^{\infty}\gamma^t\mathcal{R}_t]\), where \(\mathcal{R}_t\) reflects the SSC requirements.

    \section{Approach}\label{sec:approach}
    To address the structural incompatibility between surveillance and containment outlined in Section~\ref{sec:problem-formulation}, we propose the Dual-Stream Gated Network.
DSGN decouples the state representation process into spatial-explorative reasoning for area surveillance and tactical-exploitative coordination for target containment.
This architectural inductive bias allows each pursuer to preserve task-specific representations while a context-aware gating mechanism dynamically arbitrates the two streams to produce a unified decentralized policy.

\subsection{Observation and Input Representation}
\label{subsec:input-representation}

As defined in Section~\ref{sec:problem-formulation}, each pursuer \(n\in[N]\) receives a local observation \(o_n^t=\mathcal{O}(s^t,n)\in\Omega\).
For network processing, \(o_n^t\) is converted into a fixed-size structured input sequence \(X_n^t\), consisting of an ego token and external-entity tokens.
The ego token encodes \(x_n^t=(p_n^t,v_n^t,\psi_n^t)\), the minimum distance to locally perceived obstacles, and the evader-observation flag \(M_{\text{vis}}^t=\mathbb{I}(|\mathcal{E}_{\text{obs}}^t|>0)\).
External entities include locally perceived teammates, observed evaders in \(\mathcal{E}_{\text{obs}}^t\), visible obstacles, and surveillance nodes \(c_k\in\mathcal{C}\).
For each external entity \(\xi\) at position \(z_\xi^t\), we use the ego-centric geometric core \([z_\xi^t-p_n^t,\|z_\xi^t-p_n^t\|,\operatorname{bear}(z_\xi^t-p_n^t,\psi_n^t)]\), where \(\operatorname{bear}(\cdot)\) is the bearing relative to pursuer \(n\)'s heading.
Entity-specific attributes are appended compactly: relative velocity for dynamic entities, engagement intent for teammates, heading-orientation error for evaders, radius \(r_\ell\) for obstacles, and normalized idleness \(\bar d_k^t=\operatorname{Norm}(d_k^t)\) for surveillance nodes.
Padding and an entity-validity mask \(M_{\text{valid}}^{n,t}\) handle varying entity cardinalities.
All these features are deterministic functions of the observations defined in Section~\ref{sec:problem-formulation}, rather than additional problem assumptions.

\begin{figure}[t]
    \centering
    \includegraphics[width=\columnwidth,trim=25pt 494pt 38pt 27pt,clip]{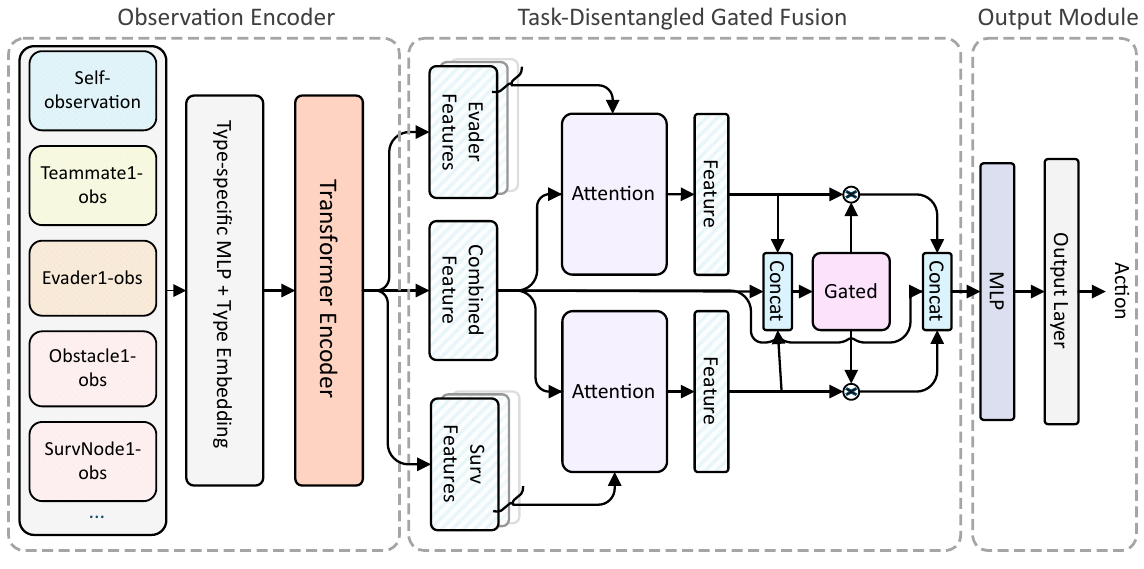}
    \caption{The proposed DSGN architecture. Type-specific observation features are encoded and fused through dual-stream attention and a vector gate to produce actions.}
    \label{fig:network-architecture}
\end{figure}

\subsection{Network Architecture}
\label{subsec:architecture}

The proposed DSGN, illustrated in Fig.~\ref{fig:network-architecture}, is an end-to-end pipeline that maps the structured input sequence \(X_n^t\) to a discrete kinematic command \(a_n^t\in\mathcal{A}\).
Instead of prematurely compressing surveillance and containment cues into a single shared representation, DSGN processes heterogeneous observations through four stages: \textit{heterogeneous feature projection}, \textit{relational context encoding}, \textit{task-disentangled gated fusion}, and \textit{distributional action inference}.
This design first aligns different entity types into a common latent space, then reasons about their spatial relations, and finally arbitrates two task-specific contexts through a learnable gate.

\subsubsection{Heterogeneous Feature Projection}

The input sequence \(X_n^t\) contains heterogeneous entities, including the ego token, teammates, observed evaders, visible obstacles, and surveillance nodes.
These entities carry different semantics and feature dimensions: for example, a surveillance node mainly reflects coverage urgency through idleness, whereas an observed evader provides target-centric geometric cues.
To avoid mixing such semantics at the raw-feature level, we use type-specific multi-layer perceptrons to project each token into a common latent space \(\mathbb{R}^d\), where \(d\) denotes the latent feature dimension.
A learnable type embedding \(E_{\text{type}}\) is added to preserve categorical identity after projection, and the validity mask \(M_{\text{valid}}^{n,t}\) prevents padded entities from contributing to attention.
This projection stage provides a unified but type-aware representation for subsequent relational reasoning.

\subsubsection{Relational Context Encoding}

After feature projection, each pursuer must reason not only about individual entities but also about their spatial and task relations.
For example, a neglected surveillance node is more informative when nearby teammates are absent, and an observed evader is more actionable when enough pursuers are already forming a partial enclosure.
We therefore employ a transformer encoder to capture high-order dependencies among the projected entity tokens.
It takes the type-aware sequence \(E_{\text{in}}^{n,t}+E_{\text{type}}\) and validity mask \(M_{\text{valid}}^{n,t}\) as input and produces \(E_{\text{ctx}}^{n,t}\), where \(E_{\text{in}}^{n,t}\) denotes the projected input sequence.
This encoder aggregates local scene geometry, teammate distribution, evader observations, obstacle constraints, and surveillance-node urgency into context-aware entity representations.
The updated ego feature \(h_{\text{ego}}^{n,t}\), extracted from the ego-token position in \(E_{\text{ctx}}^{n,t}\), serves as the central query for dual-stream decision-making.

\subsubsection{Task-Disentangled Gated Fusion}

Given the surveillance--containment conflict discussed in Section~\ref{sec:problem-formulation}, DSGN uses two specialized attention streams to maintain task-specific contexts before fusion.

\textbf{Spatial-Explorative Surveillance Stream:}
This stream focuses on coverage-oriented reasoning.
It uses the ego feature \(h_{\text{ego}}^{n,t}\) to query encoded surveillance-node features, producing a surveillance context \(h_{\text{sur}}^{n,t}\).
Because surveillance-node tokens contain idleness information and spatial relations to the current team distribution, \(h_{\text{sur}}^{n,t}\) highlights neglected regions and encourages dispersion toward coverage gaps.
\textbf{Tactical-Exploitative Containment Stream:}
This stream focuses on target-centered coordination.
It uses \(h_{\text{ego}}^{n,t}\) to query encoded observed-evader and teammate features, producing a containment context \(h_{\text{con}}^{n,t}\).
This representation captures relative target geometry, teammate support, and local formation requirements.
When no evader is observed, the stream still processes teammate configurations, which helps preserve tactical readiness without forcing premature containment behavior.
\textbf{Dynamic Arbitration via Vector-wise Gating:}
To balance the two task contexts, DSGN generates a vector gate \(g_n^t\in(0,1)^d\) by applying the sigmoid function \(\sigma(\cdot)\) to \(\operatorname{MLP}_{\text{gate}}\) of the concatenated features \(h_{\text{ego}}^{n,t}\oplus h_{\text{sur}}^{n,t}\oplus h_{\text{con}}^{n,t}\), where \(\oplus\) denotes concatenation.
Each gate dimension modulates one latent channel.
The fused representation \(h_{\text{fused}}^{n,t}\) is the projection of the concatenated features \(h_{\text{ego}}^{n,t}\), \(g_n^t\odot h_{\text{con}}^{n,t}\), and \((1-g_n^t)\odot h_{\text{sur}}^{n,t}\), where \(\odot\) denotes the Hadamard product.
Large gate values emphasize containment-related channels, whereas small values emphasize surveillance-related channels.
Unlike a scalar switch, this vector-wise gate allows different latent dimensions to encode different aspects of the current situation, such as coverage urgency, target proximity, teammate support, and obstacle risk.
As a result, the policy can maintain surveillance behavior while gradually committing to containment when target evidence becomes reliable.

\subsubsection{Distributional Action Inference}

The fused representation \(h_{\text{fused}}^{n,t}\) is mapped to kinematic decisions using an implicit quantile network (IQN)~\cite{dabney2018implicit}.
Instead of estimating only a scalar action value, IQN approximates the return distribution \(Z_\theta(o_n^t,a)\) for each action \(a\in\mathcal{A}\).
This is useful in SSC because partial observability and multi-robot interactions can make the outcome of the same action highly variable across local contexts.

\subsection{Multi-Objective Reward Design}
\label{subsec:reward}

Section~\ref{sec:problem-formulation} defines the SSC objective as maintaining persistent surveillance while satisfying the containment condition in Eq.~\eqref{eq:capture_cond}.
Since this objective cannot be directly used as a reinforcement-learning signal, we design \(\mathcal{R}_t\) as a shaped reward that reflects surveillance, containment, safety, and efficiency requirements.
This design converts the task-level SSC objective into dense learning signals for decentralized policy optimization.
For surveillance, pursuer \(n\) receives the arrival reward \(r_{\text{arr}}^{n,t}=\psi_{\text{arr}}\sum_{k\in[K]}\bar d_k^t\mathbb{I}_{\text{visit}}^{n,k,t}\) when visiting urgent nodes, where \(\psi_{\text{arr}}>0\), \(\bar d_k^t=\operatorname{Norm}(d_k^t)\), and \(\mathbb{I}_{\text{visit}}^{n,k,t}=\mathbb{I}(\|p_n^t-c_k\|\le R_{\text{visit}})\).
Because arrival rewards are sparse, we add a potential-based shaping term toward the most urgent node \(k^*(t)=\arg\max_{k\in[K]}d_k^t\).
The shaping reward is \(r_{\text{shape}}^{n,t}=\psi_{\text{shape}}(\|p_n^{t-1}-c_{k^*(t)}\|-\|p_n^t-c_{k^*(t)}\|)\), where \(\psi_{\text{shape}}>0\).
This follows policy-invariant reward shaping~\cite{ng1999policy} to get a smooth gradient toward neglected surveillance regions.

For containment guidance, let \(d_{\min}^{n,t}=\min_{m\in\mathcal{E}_{\text{obs}}^t}\|p_n^t-e_m^t\|\) denote the distance to the closest observed evader when \(\mathcal{E}_{\text{obs}}^t\neq\emptyset\).
To reduce overcrowding around the same target, each evader admits only the \(N_{\text{eng}}\) closest pursuers into its engagement set, where \(N_{\text{eng}}\) is a fixed parameter.
The engagement indicator \(\mathbb{I}_{\text{eng}}^{n,t}\) is one if pursuer \(n\) is non-redundant and \(d_{\min}^{n,t}<d_{\text{react}}\), with \(\mathbb{I}_{\text{sur}}^{n,t}=1-\mathbb{I}_{\text{eng}}^{n,t}\).
When no evader is observed, \(\mathbb{I}_{\text{eng}}^{n,t}=0\), \(\mathbb{I}_{\text{sur}}^{n,t}=1\), and \(r_{\text{guide}}^{n,t}=0\).
If \(d_{\min}^{n,t}<d_{\text{buf}}\), the containment guidance reward is \(r_{\text{guide}}^{n,t}=-\psi_{\text{rep}}(d_{\text{buf}}-d_{\min}^{n,t})\); otherwise, it is \(r_{\text{guide}}^{n,t}=\psi_{\text{app}}(d_{\min}^{n,t-1}-d_{\min}^{n,t})\), where \(d_{\text{react}},d_{\text{buf}},\psi_{\text{rep}},\psi_{\text{app}}>0\).
The first term prevents unsafe proximity to evaders, while the second provides a continuous approach gradient toward containment formation.

Finally, \(r_{\text{col}}^{n,t}\) penalizes dangerous proximity to teammates or obstacles, \(r_{\text{con}}^{n,t}\) rewards satisfying Eq.~\eqref{eq:capture_cond}, and \(r_{\text{win}}^{n,t}\) rewards full task completion.
We also include a small step penalty \(r_{\text{time}}^{n,t}<0\) to encourage efficient task completion.
The total reward \(r_{\text{total}}^{n,t}\) sums \(r_{\text{arr}}^{n,t}\), the gated terms \(\mathbb{I}_{\text{sur}}^{n,t}r_{\text{shape}}^{n,t}\) and \(\mathbb{I}_{\text{eng}}^{n,t}r_{\text{guide}}^{n,t}\), and \(r_{\text{col}}^{n,t}+r_{\text{con}}^{n,t}+r_{\text{win}}^{n,t}+r_{\text{time}}^{n,t}\).

\subsection{Training Strategy via Probabilistic Prior Injection}
\label{subsec:training}

While DSGN is solver-agnostic, we employ IQN to model distributional returns.
Optimizing the vector gate \(g_n^t\in(0,1)^d\) solely via extrinsic rewards can yield ambiguous weights, e.g., \(g_n^t\approx0.5\mathbf{1}_d\), where \(\mathbf{1}_d\) is the \(d\)-dimensional all-one vector.
We therefore regularize each channel \(g_{n,j}^t\), \(j=1,\dots,d\), with a state-dependent Beta prior whose parameters depend on \(M_{\text{vis}}^t\): \((\alpha,\beta)=(1,4)\) if \(M_{\text{vis}}^t=0\), and \((\alpha,\beta)=(0.5,0.5)\) if \(M_{\text{vis}}^t=1\).
The resulting negative log-likelihood regularizer \(\mathcal{L}_{\text{reg}}(g_n^t,M_{\text{vis}}^t)\) is the sum over \(j=1,\ldots,d\) of \((\alpha(M_{\text{vis}}^t)-1)\ln g_{n,j}^t\) and \((\beta(M_{\text{vis}}^t)-1)\ln(1-g_{n,j}^t)\), multiplied by \(-1/d\).

\begin{figure}[t]
    \centering
    \vspace{1.8mm}
    \resizebox{\columnwidth}{!}{%
    \begin{tikzpicture}
    \begin{groupplot}[
        group style={
            group size=2 by 1,
            horizontal sep=1.45cm
        },
        width=0.455\textwidth,
        height=5.55cm,
        xmin=0, xmax=1,
        ymin=0, ymax=4.5,
        xtick={0,0.2,0.4,0.6,0.8,1},
        ytick={0,1,2,3,4},
        axis lines=box,
        axis line style={black, line width=0.6pt},
        tick align=outside,
        samples=500,
        domain=0.001:0.999,
        xlabel={Per-channel Gate $g_{n,j}^t$},
        title style={align=center, font=\fontsize{12}{13}\selectfont},
        tick label style={font=\normalsize},
        label style={font=\normalsize},
        x label style={font=\fontsize{12}{13}\selectfont},
        y label style={font=\fontsize{12}{13}\selectfont},
    ]

    \nextgroupplot[
        ylabel={Density},
        title={\textbf{(a) Sparsity Prior} ($M_{\text{vis}}^t=0$)\\Constraint: Suppress Noise},
    ]

    \addplot[
        draw=none,
        fill=COLORSPARSITY,
        fill opacity=0.28,
        domain=0:1
    ] {4*(1-x)^3} \closedcycle;

    \addplot[
        color=COLORSPARSITY,
        line width=1.7pt,
        domain=0:1
    ] {4*(1-x)^3};

    \node[
        align=center,
        font=\normalsize,
        fill=white,
        fill opacity=0.85,
        text opacity=1,
        inner sep=2pt
    ] at (axis cs:0.52,3.55) {$\alpha=1,\ \beta=4$};

    \draw[->, line width=1.15pt, color=COLORARROW]
        (axis cs:0.40,1.32) -- (axis cs:0.20,1.32);
    \draw[->, line width=1.15pt, color=COLORARROW]
        (axis cs:0.64,0.45) -- (axis cs:0.46,0.45);
    \draw[->, line width=1.15pt, color=COLORARROW]
        (axis cs:0.90,0.20) -- (axis cs:0.70,0.20);

    \node[
        align=center,
        text=COLORARROW,
        font=\normalsize,
        fill=white,
        fill opacity=0.82,
        text opacity=1,
        inner sep=2pt
    ] at (axis cs:0.53,2.58) {Gradient Field pushes $g_{n,j}^t \to 0$};

    \nextgroupplot[
        title={\textbf{(b) Decisiveness Prior} ($M_{\text{vis}}^t=1$)\\Constraint: Binary Decision},
    ]

    \addplot[
        draw=none,
        fill=COLORDECISIVE,
        fill opacity=0.28,
        domain=0.001:0.999
    ] {1/(pi*sqrt(x*(1-x)))} \closedcycle;

    \addplot[
        color=COLORDECISIVE,
        line width=1.7pt,
        domain=0.001:0.999
    ] {1/(pi*sqrt(x*(1-x)))};

    \node[
        align=center,
        font=\normalsize,
        fill=white,
        fill opacity=0.85,
        text opacity=1,
        inner sep=2pt
    ] at (axis cs:0.50,3.55) {$\alpha=\beta=0.5$};

    \draw[->, line width=1.15pt, color=COLORARROW]
        (axis cs:0.40,1.5) -- (axis cs:0.10,1.5);
    \draw[->, line width=1.15pt, color=COLORARROW]
        (axis cs:0.60,1.5) -- (axis cs:0.90,1.5);

    \node[
        align=center,
        fill=white,
        fill opacity=0.90,
        text opacity=1,
        inner sep=2pt,
        font=\normalsize
    ] at (axis cs:0.50,2.20)
    {\textbf{Unstable Equilibrium}\\at $g_{n,j}^t=0.5$};

    \end{groupplot}
    \end{tikzpicture}%
    }
    \caption{State-dependent Beta priors for gate regularization.}
    \label{fig:beta_priors}
    \vspace{-0.2cm}
\end{figure}
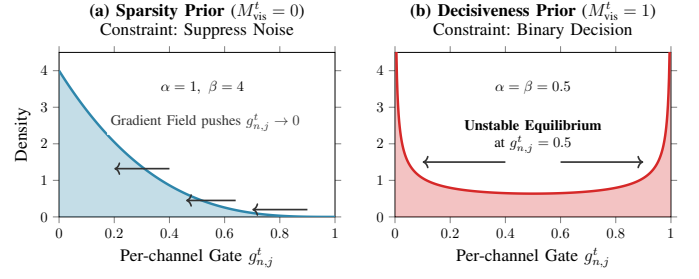

Fig.~\ref{fig:beta_priors} illustrates the state-dependent priors applied to each gate channel.
When no evader is observed, \((\alpha,\beta)=(1,4)\) drives \(g_{n,j}^t\to0\) and suppresses noisy containment activation.
When evaders are observed, \(\alpha=\beta=0.5\) penalizes ambiguous gates near \(0.5\) and encourages decisive arbitration between surveillance and containment.

To prevent the prior from dominating early representation learning, we linearly increase its coefficient \(\lambda_t\) after a warm-up phase.
The final objective \(\mathcal{L}_{\text{total}}\) sums the average quantile Huber loss \(N_q^{-1}\sum_{q=1}^{N_q}\rho_{\tau_q}^{\kappa}(\delta_{\tau_q})\)~\cite{dabney2018implicit} and the weighted gate regularizer \(\lambda_t\mathcal{L}_{\text{reg}}(g_n^t,M_{\text{vis}}^t)\), where \(\delta_{\tau_q}\) is the temporal-difference error at sampled quantile \(\tau_q\sim U([0,1])\), \(N_q\) is the number of quantile samples, and \(\kappa\) is the Huber threshold.

    \section{Experiments}\label{sec:experiments}
\vspace{1.5mm}
\begin{table}[t]
    \centering
    \scriptsize
    \setlength{\tabcolsep}{0pt}
    \setlength{\aboverulesep}{1pt}
    \setlength{\belowrulesep}{1pt}
    \caption{\textbf{Progressive Curriculum Schedule}}
    \label{tab:curriculum_schedule}
    \begin{tabular*}{\columnwidth}{@{\extracolsep{\fill}} l c c c @{}}
        \toprule
        \textbf{Training Steps ($10^6$)} & \textbf{Pursuers} & \textbf{Evaders} & \textbf{Obstacles} \\
        \midrule
        \renewcommand{\arraystretch}{0.95}
        \(0 \leq T_{\text{train}} < 2\)           & 3                 & 1                & 0                  \\
        \(2 \leq T_{\text{train}} < 4\)           & 4                 & 1                & 1                  \\
        \(4 \leq T_{\text{train}} < 5\)           & 7                 & 2                & 2                  \\
        \(5 \leq T_{\text{train}} < 6\)           & 11                & 3                & 4                  \\
        \(6 \leq T_{\text{train}} \leq 7\)        & 15                & 4                & 6                  \\
        \bottomrule
    \end{tabular*}
    \vspace{-6mm}
\end{table}

\begin{figure*}[t]
    \centering
    \vspace{1.8mm}
    \includegraphics[width=0.85\textwidth]{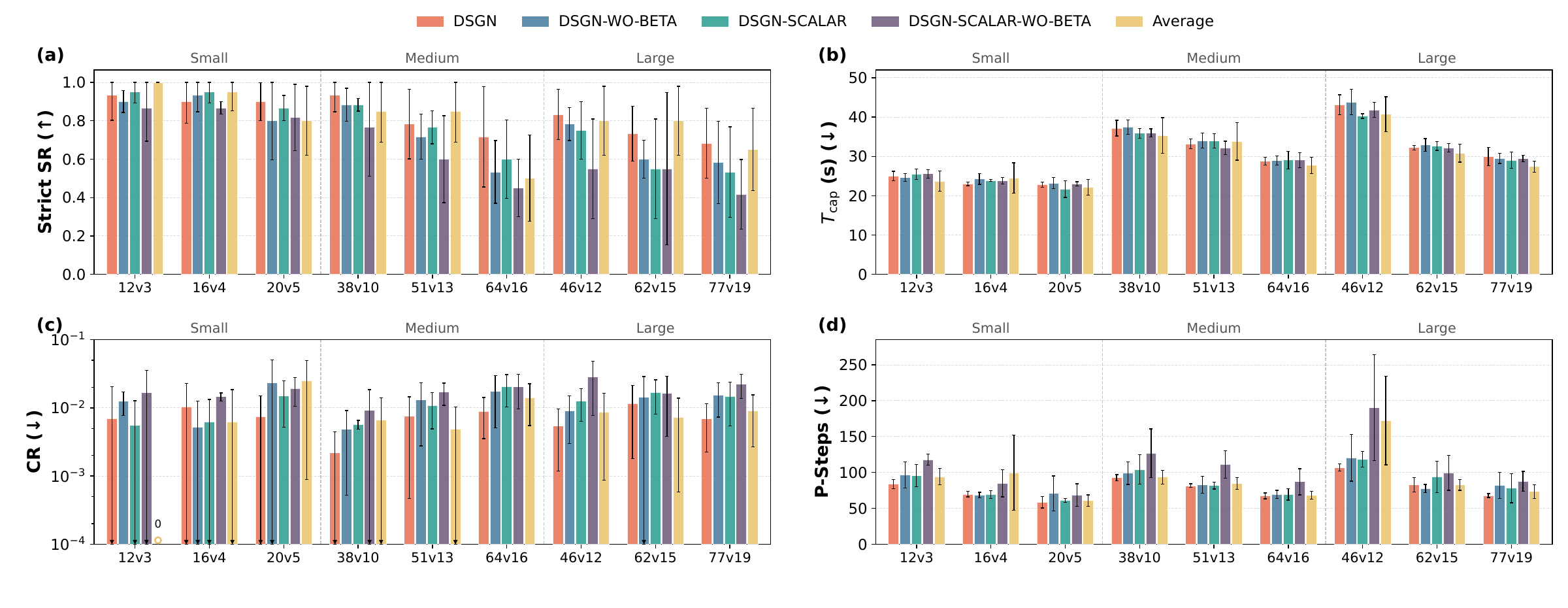}
    \caption{\textcolor{black}{\textbf{Arbitration Variants across Nine Scale--Density Scenarios.} \textbf{(a)} Strict SR; \textbf{(b)} \(T_{\mathrm{cap}}\) (s); \textbf{(c)} CR (log scale); \textbf{(d)} P-Steps. Error bars show 95\% confidence intervals across 5 independent training seeds, with 40 trials per seed and scenario.}}
    \label{fig:exp_results}
    \vspace{-5mm}
\end{figure*}

\subsection{Training Setup}\label{subsec:training-setup}

\textcolor{black}{
We extend the open-source multi-robot navigation simulator described in~\cite{lin2024decentralized} by retaining its kinematic, sensing, and collision models and adding surveillance nodes, reactive evaders, and multi-target containment.}
The workspace is discretized into a uniform rectangular grid, and one surveillance node \(c_k\) is placed at the center of each grid cell for persistent surveillance and idleness computation.
For an environment of size \(W\times H\), the maximum grid-cell side length is \(L_{\max}=\eta R_{\text{sense}}\sqrt{2}\), ensuring that half the diagonal of each cell is no larger than \(\eta R_{\text{sense}}\).
With \(R_{\text{sense}}=20\,\text{m}\) and \(\eta=0.95\), this gives \(L_{\max}\approx26.9\,\text{m}\).
The numbers of grid cells along the two axes are \(N_x=\lceil W/L_{\max}\rceil\) and \(N_y=\lceil H/L_{\max}\rceil\), with cell dimensions \(\Delta_x=W/N_x\) and \(\Delta_y=H/N_y\).

During training, each agent's input includes at most the nearest 5 obstacles, 5 teammates, 8 evaders, and 16 surveillance nodes; unused entries are zero-padded.
Successful containment follows Eq.~\eqref{eq:capture_cond} with \(\lambda_{\text{uni}}=3\).
A training episode terminates when all evaders are contained, fewer than three pursuers remain active, or the episode exceeds 3000 time steps.
To ease learning in the early stages, training follows a progressive curriculum that gradually increases the number of entities, as shown in Table~\ref{tab:curriculum_schedule}.

\subsection{Evaluation Setup}\label{subsec:evaluation-setup}

\textcolor{black}{Without retraining, each policy is tested across nine scale--density configurations (Table~\ref{tab:evaluation_schedule}), ranging from 12v3 (12 pursuers versus 3 evaders) to 77v19. The largest evaluation setting, 77v19, exceeds the largest curriculum training setting, 15v4.}
Within each density level, the environment size varies with team size to maintain approximately constant mobile-entity density across map scales.
This design helps separate the effects of increasing team and map sizes from changes in overall crowding.
\textcolor{black}{All experiments use a consistent rectangular outer boundary.}
\textcolor{black}{Except for the separately reported action-granularity test in Section~\ref{subsec:action_granularity}, each method and arbitration variant is evaluated over 40 trials per random seed, and across 5 independent random seeds in every task scenario. Initial poses of pursuers, evaders, and obstacles are randomized in each trial. For these comparisons, results are reported as seed-level means across seeds with 95\% confidence intervals. These tests assess structural generalization under shared simulator dynamics. The evaluation horizon is 1000 time steps, corresponding to 500s of simulated time.}

We impose kinematic asymmetry so that successful containment requires cooperation.
\textcolor{black}{Following the discrete kinematic command setting in~\cite{lin2024decentralized}, pursuers are bounded by \(v_{\max}=3.0\) m/s and use a \(3\times3\) action space (\(a_{\text{acc}}\in\{0,\pm0.4\}\,\text{m/s}^2\), \(\omega\in\{0,\pm\pi/6\}\,\text{rad/s}\)). These actions form a platform-agnostic high-level command interface; related learned pursuit policies have been deployed in real-world settings~\cite{zhang2022multi}.}
Evaders are guided by a reactive artificial potential field (APF)~\cite{khatib1986real} and use \(v_{\max}=3.5\) m/s with a finer rotational spectrum (\(\omega \in \{0, \pm \pi/12, \pm \pi/6\}\) rad/s) under the same acceleration limits.
This kinematic difference precludes naive individual pursuit.

\begin{table}[t!]
    \centering
    \vspace{1.8mm}
    \footnotesize
    \color{black}
    \caption{\textcolor{black}{\textbf{Configurations for Cross-Scale Evaluation}}}
    \label{tab:evaluation_schedule}

    \setlength{\aboverulesep}{1pt}
    \setlength{\belowrulesep}{1pt}
    \setlength{\tabcolsep}{2.2pt}
    \begin{tabular}{@{}llcccc@{}}
        \toprule
        \makecell{\textbf{Map} \\ \textbf{Scale}}
        & \makecell{\textbf{Region Size} \\ \textbf{(m$\times$m)}}
        & \textbf{Pursuers} & \textbf{Evaders} & \textbf{Obstacles}
        & \makecell{{\scriptsize\textbf{Density Level}} \\ {\scriptsize$\boldsymbol{\rho_m}$ ($10^{-3}$ m$^{-2}$)}} \\
        \midrule
        \multirow{3}{*}{Small}
        & \multirow{3}{*}{$112 \times 112$} & 12 & 3 & 3 & {\scriptsize Low (1.196)} \\
        &                                      & 16 & 4 & 3 & {\scriptsize Medium (1.594)} \\
        &                                      & 20 & 5 & 3 & {\scriptsize High (1.993)} \\
        \midrule
        \multirow{3}{*}{Medium}
        & \multirow{3}{*}{$200 \times 200$} & 38 & 10 & 10 & {\scriptsize Low (1.200)} \\
        &                                      & 51 & 13 & 10 & {\scriptsize Medium (1.600)} \\
        &                                      & 64 & 16 & 10 & {\scriptsize High (2.000)} \\
        \midrule
        \multirow{3}{*}{Large}
        & \multirow{3}{*}{$220 \times 220$} & 46 & 12 & 12 & {\scriptsize Low (1.198)} \\
        &                                      & 62 & 15 & 12 & {\scriptsize Medium (1.591)} \\
        &                                      & 77 & 19 & 12 & {\scriptsize High (1.983)} \\
        \bottomrule
    \end{tabular}
    \vspace{0.6mm}
    \begin{minipage}{\columnwidth}
        \scriptsize
        \emph{Note:} $\rho_m=(N+M)/(WH)$ denotes the mobile-entity density.
    \end{minipage}
    \vspace{-6mm}
\end{table}

We evaluate cross-scale performance using five metrics.
\textit{Strict Success Rate} (SR) is the fraction of trials in which all evaders are contained without collisions.
\textcolor{black}{\textit{Containment Time} (\(T_{\mathrm{cap}}\)) measures the time from episode start to containment completion, conditioned on episodes that complete containment.}
\textcolor{black}{\textit{Collision Ratio} (CR) is computed per episode as the number of pursuers that experience at least one collision with an obstacle or teammate, divided by the total number of pursuers; repeated collisions by the same pursuer are counted once.}
\textit{Containment Angle Variance} (Var) is computed when an individual evader is contained and measures the angular uniformity of the containment formation.
\textcolor{black}{\textit{Patrol Completion Steps} (P-Steps) measures the steps from containment completion to final surveillance coverage, conditioned on completion of both stages.}

\subsection{Baseline Methods}\label{subsec:baseline-methods}

We select representative baselines relevant to decentralized decision-making, multi-agent motion planning, and reinforcement learning, and adapt them to the SSC setting.
We compare DSGN with TransIQN~\cite{zhang2025terl}, IQN~\cite{dabney2018implicit}, Deep Q-Network (DQN)~\cite{mnih2015human}, and a reactive Heuristic baseline based on APF~\cite{khatib1986real} under identical settings.
\textcolor{black}{For the main comparisons, all learned methods use the same shaped reward (Section~\ref{subsec:reward}), progressive curriculum (Table~\ref{tab:curriculum_schedule}), action space, and 7M steps training budget per seed.}
The TransIQN baseline uses the same transformer backbone as DSGN but represents surveillance and containment in a single ego-token feature, providing a direct comparison between shared and task-specific relational representations.
To control for model capacity, its parameter count is matched to that of DSGN using a residual expansion block.
IQN and DQN serve as representative value-based learning baselines, but both compress environmental features through average pooling and therefore do not perform explicit relational reasoning.
The Heuristic baseline uses manually designed attractive and repulsive fields as a reactive policy without learning.

\subsection{Results and Discussion}\label{subsec:results-and-discussion}

\textcolor{black}{Fig.~\ref{fig:exp_results} compares DSGN, its three gate and prior ablations, and the fixed equal-weight Average variant across the nine configurations in Table~\ref{tab:evaluation_schedule}; Table~\ref{tab:ablation_gating} summarizes the results by scale.}

\textcolor{black}{\emph{Task Completion}: DSGN has the highest or tied-highest scale-averaged Strict SR at all three scales in Fig.~\ref{fig:exp_results}(a), despite not leading in every configuration. This supports more reliable task completion through separate surveillance and containment streams.}

\textcolor{black}{\emph{Containment Efficiency}: In Fig.~\ref{fig:exp_results}(b), DSGN's scale-averaged \(T_{\mathrm{cap}}\) exceeds the fastest variant by only 0.1--2.1~s. Its short times suggest that relational attention aids spatial assignment around evaders, while dual-stream processing adds little execution delay.}

\textcolor{black}{\emph{Safety}: DSGN has the lowest scale-averaged CR at all three scales in Fig.~\ref{fig:exp_results}(c), although the best variant varies by configuration. This suggests that disentangled task processing reduces interference between surveillance and containment behaviors.}

\textcolor{black}{\emph{Post-Containment Surveillance}: DSGN has the lowest scale-averaged P-Steps at all three scales in Fig.~\ref{fig:exp_results}(d), despite some configuration-level exceptions. This suggests that its surveillance stream supports prompt coverage completion after containment.}

\textcolor{black}{Together, Fig.~\ref{fig:exp_results} and Table~\ref{tab:ablation_transition} suggest that task-specific streams complement relational reasoning: unlike a shared stream that may mix surveillance and containment cues, DSGN addresses this issue by separating the two objectives into specialized streams and coordinating them through context-aware gating. This design supports safer and more consistent multi-robot coordination as the number of interacting agents increases.}

\begin{table}[t]
    \centering
    \scriptsize
    \caption{\textcolor{black}{\textbf{Mode Switching and Pure Surveillance Performance.} Each entry reports SR / Time (s), with 95\% confidence intervals in brackets.}}
    \label{tab:ablation_transition}

    \setlength{\tabcolsep}{2.2pt}

    {\color{black}
    \newcommand{\tabpair}[6]{\makecell{#1 / #4\\{\tiny [#2, #3] / [#5, #6]}}}
    \textit{(a) Task Switching} \hfill \((\text{SR}\uparrow / \text{Time}\downarrow)\)
    \vspace{0.5mm}

    \resizebox{0.95\columnwidth}{!}{%
    \begin{tabular}{@{}lccc@{}}
        \toprule
        \textbf{Method} & \textbf{3v1} & \textbf{6v2} & \textbf{9v3} \\
        \midrule
        Heuristic
        & \tabpair{0.000}{0.000}{0.000}{488.4}{465.0}{511.9}
        & \tabpair{0.000}{0.000}{0.000}{500.0}{500.0}{500.0}
        & \tabpair{0.000}{0.000}{0.000}{500.0}{500.0}{500.0} \\
        DQN
        & \tabpair{0.000}{0.000}{0.000}{500.0}{500.0}{500.0}
        & \tabpair{0.000}{0.000}{0.000}{500.0}{500.0}{500.0}
        & \tabpair{0.000}{0.000}{0.000}{500.0}{500.0}{500.0} \\
        IQN
        & \tabpair{0.000}{0.000}{0.000}{451.3}{415.4}{487.1}
        & \tabpair{0.950}{0.879}{1.000}{169.5}{155.4}{183.6}
        & \tabpair{0.785}{0.650}{0.920}{112.5}{106.7}{118.3} \\
        TransIQN
        & \tabpair{0.025}{0.000}{0.076}{455.0}{413.8}{496.2}
        & \tabpair{0.840}{0.724}{0.956}{190.8}{160.6}{220.9}
        & \tabpair{0.725}{0.580}{0.870}{103.9}{84.8}{122.9} \\
        \midrule
        Random
        & \tabpair{0.000}{0.000}{0.000}{467.2}{429.6}{504.8}
        & \tabpair{0.075}{0.000}{0.160}{475.8}{450.3}{501.4}
        & \tabpair{0.300}{0.152}{0.448}{405.7}{359.3}{452.1} \\
        Average
        & \tabpair{0.025}{0.000}{0.076}{487.7}{463.9}{511.5}
        & \tabpair{0.665}{0.513}{0.817}{337.7}{289.3}{386.0}
        & \tabpair{0.820}{0.690}{0.950}{146.6}{106.6}{186.6} \\
        Rule-based
        & \tabpair{\textbf{1.000}}{1.000}{1.000}{175.5}{164.3}{186.7}
        & \tabpair{0.925}{0.840}{1.000}{109.4}{102.8}{115.9}
        & \tabpair{0.650}{0.496}{0.804}{91.3}{85.1}{97.5} \\
        \midrule
        DSGN-WO-BETA
        & \tabpair{0.165}{0.042}{0.288}{473.7}{451.8}{495.5}
        & \tabpair{\textbf{0.975}}{0.924}{1.000}{128.1}{118.7}{137.4}
        & \tabpair{0.875}{0.768}{0.982}{86.1}{81.3}{90.8} \\
        DSGN-SCALAR
        & \tabpair{0.975}{0.924}{1.000}{225.1}{204.0}{246.1}
        & \tabpair{\textbf{0.975}}{0.924}{1.000}{122.7}{99.4}{146.1}
        & \tabpair{0.850}{0.734}{0.966}{84.6}{78.0}{91.2} \\
        DSGN-SCALAR-WO-BETA
        & \tabpair{0.925}{0.840}{1.000}{317.6}{287.2}{347.9}
        & \tabpair{0.925}{0.840}{1.000}{138.6}{125.2}{152.0}
        & \tabpair{0.810}{0.680}{0.940}{93.9}{82.1}{105.7} \\
        \midrule
        \textbf{DSGN}
        & \tabpair{\textbf{1.000}}{1.000}{1.000}{\textbf{162.2}}{150.6}{173.7}
        & \tabpair{0.950}{0.879}{1.000}{\textbf{98.0}}{93.0}{103.0}
        & \tabpair{\textbf{0.925}}{0.840}{1.000}{\textbf{76.7}}{72.2}{81.3} \\
        \bottomrule
    \end{tabular}
    }

    \vspace{1.2mm}
    \textit{(b) Pure Surveillance, where \(N\) denotes the number of pursuers} \hfill \((\text{SR}\uparrow / \text{Time}\downarrow)\)
    \vspace{0.5mm}

    \resizebox{0.95\columnwidth}{!}{%
    \begin{tabular}{@{}lccc@{}}
        \toprule
        \textbf{Method} & \textbf{$N=3$} & \textbf{$N=4$} & \textbf{$N=5$} \\
        \midrule
        Heuristic
        & \tabpair{0.000}{0.000}{0.000}{500.0}{500.0}{500.0}
        & \tabpair{0.000}{0.000}{0.000}{500.0}{500.0}{500.0}
        & \tabpair{0.000}{0.000}{0.000}{500.0}{500.0}{500.0} \\
        DQN
        & \tabpair{0.000}{0.000}{0.000}{500.0}{500.0}{500.0}
        & \tabpair{0.000}{0.000}{0.000}{500.0}{500.0}{500.0}
        & \tabpair{0.075}{0.000}{0.160}{494.1}{486.9}{501.2} \\
        IQN
        & \tabpair{0.000}{0.000}{0.000}{444.6}{410.2}{479.0}
        & \tabpair{0.385}{0.228}{0.542}{432.9}{400.6}{465.2}
        & \tabpair{0.900}{0.803}{0.997}{288.5}{253.6}{323.4} \\
        TransIQN
        & \tabpair{0.175}{0.052}{0.298}{436.1}{400.3}{472.0}
        & \tabpair{0.715}{0.570}{0.860}{294.2}{244.1}{344.4}
        & \tabpair{\textbf{1.000}}{1.000}{1.000}{131.7}{112.4}{151.1} \\
        \midrule
        Random
        & \tabpair{\textbf{1.000}}{1.000}{1.000}{130.7}{122.0}{139.3}
        & \tabpair{0.975}{0.924}{1.000}{104.0}{97.0}{110.9}
        & \tabpair{\textbf{1.000}}{1.000}{1.000}{\textbf{82.5}}{77.0}{88.0} \\
        Average
        & \tabpair{0.250}{0.110}{0.390}{455.3}{425.6}{485.0}
        & \tabpair{0.620}{0.451}{0.769}{385.8}{344.1}{427.5}
        & \tabpair{0.875}{0.768}{0.982}{245.9}{204.6}{287.1} \\
        Rule-based
        & \tabpair{\textbf{1.000}}{1.000}{1.000}{137.7}{129.0}{146.3}
        & \tabpair{\textbf{1.000}}{1.000}{1.000}{105.2}{99.1}{111.2}
        & \tabpair{\textbf{1.000}}{1.000}{1.000}{84.8}{80.5}{89.0} \\
        \midrule
        DSGN-WO-BETA
        & \tabpair{0.660}{0.506}{0.814}{379.5}{340.8}{418.1}
        & \tabpair{0.975}{0.924}{1.000}{200.7}{170.4}{231.1}
        & \tabpair{\textbf{1.000}}{1.000}{1.000}{108.9}{99.4}{118.5} \\
        DSGN-SCALAR
        & \tabpair{0.975}{0.924}{1.000}{177.4}{166.4}{188.5}
        & \tabpair{\textbf{1.000}}{1.000}{1.000}{167.8}{148.5}{187.0}
        & \tabpair{\textbf{1.000}}{1.000}{1.000}{110.8}{98.4}{123.2} \\
        DSGN-SCALAR-WO-BETA
        & \tabpair{0.925}{0.840}{1.000}{287.8}{252.1}{323.4}
        & \tabpair{0.950}{0.879}{1.000}{137.2}{113.4}{161.0}
        & \tabpair{0.975}{0.924}{1.000}{102.9}{81.8}{123.9} \\
        \midrule
        \textbf{DSGN}
        & \tabpair{\textbf{1.000}}{1.000}{1.000}{\textbf{124.6}}{118.1}{131.1}
        & \tabpair{\textbf{1.000}}{1.000}{1.000}{\textbf{100.3}}{95.6}{105.1}
        & \tabpair{\textbf{1.000}}{1.000}{1.000}{83.3}{77.9}{88.7} \\
        \bottomrule
    \end{tabular}
    }

    \vspace{0.6mm}
    \begin{minipage}{\columnwidth}
        \scriptsize \emph{Note:} Task-switching success requires complete coverage and containment. Time is mean episode duration over all trials, capped at 500~s per trial; confidence intervals are not clipped.
    \end{minipage}
    }

\end{table}

\subsection{Ablation Study}\label{subsec:ablation-study}

\textcolor{black}{To analyze task arbitration and mode switching, we conduct structural ablations and phase-specific evaluations (Tables~\ref{tab:ablation_transition}--\ref{tab:temporal_switching_metrics} and Table~\ref{tab:ablation_gating}); we also assess sensitivity to search-region geometry and action granularity (Tables~\ref{tab:geometry_ablation} and~\ref{tab:action_granularity}).}
Random replaces the learned gate with a random gate; Average assigns equal weights to the two streams; Rule-based uses a hand-crafted spatial rule for arbitration; DSGN-WO-BETA removes the Beta-prior regularization; DSGN-SCALAR replaces vector-wise gating with a single scalar gate; and DSGN-SCALAR-WO-BETA combines scalar gating without the Beta prior.

\subsubsection{Dynamic Gating and Probabilistic Priors}

\textcolor{black}{Due to space constraints, Table~\ref{tab:ablation_gating_large} presents only the Large-scale results. The complete comparison across Small, Medium, and Large scales is provided in Table~\ref{tab:ablation_gating} in Appendix~\ref{app:additional-results}; each scale aggregates the Low, Medium, and High density configurations in Table~\ref{tab:evaluation_schedule}.}
\textcolor{black}{Removing either the Beta prior or vector-wise gating worsens scale-averaged SR, CR, Var, and P-Steps at all three scales (Table~\ref{tab:ablation_gating}). This supports both components' roles in reliable task arbitration, while Average retains a modest \(T_{\mathrm{cap}}\) advantage.}

\begin{table}[t]
    \centering
    \caption{\textcolor{black}{\textbf{Large-Scale Arbitration Variants (46v12--77v19).} Each cell reports the mean with a 95\% confidence interval in brackets.}}
    \label{tab:ablation_gating_large}

    {\color{black}
    \scriptsize
    \setlength{\tabcolsep}{2pt}
    \renewcommand{\arraystretch}{1.08}
    \resizebox{0.95\columnwidth}{!}{%
    \begin{tabular}{@{}lccccc@{}}
        \toprule
        \textbf{Variant} & SR $\uparrow$ & $T_{\mathrm{cap}}$ (s) $\downarrow$ & CR $\downarrow$ & Var $\downarrow$ & P-Steps $\downarrow$ \\
        \midrule
        Random & \makecell{0.000\\{\scriptsize [0.000, 0.000]}} & \makecell{380.9\\{\scriptsize [303.6, 458.1]}} & \makecell{0.1403\\{\scriptsize [0.1173, 0.1634]}} & \makecell{2.392\\{\scriptsize [2.276, 2.507]}} & \makecell{124.8\\{\scriptsize [113.8, 135.8]}} \\
        Average & \makecell{\textbf{0.750}\\{\scriptsize [0.640, 0.860]}} & \makecell{\textbf{33.0}\\{\scriptsize [30.8, 35.2]}} & \makecell{0.0083\\{\scriptsize [0.0044, 0.0123]}} & \makecell{0.836\\{\scriptsize [0.752, 0.919]}} & \makecell{109.4\\{\scriptsize [85.9, 133.0]}} \\
        Rule-based & \makecell{0.400\\{\scriptsize [0.275, 0.525]}} & \makecell{47.6\\{\scriptsize [43.5, 51.6]}} & \makecell{0.0320\\{\scriptsize [0.0219, 0.0422]}} & \makecell{1.056\\{\scriptsize [0.972, 1.140]}} & \makecell{113.6\\{\scriptsize [105.2, 122.0]}} \\
        \midrule
        DSGN-WO-BETA & \makecell{0.656\\{\scriptsize [0.540, 0.771]}} & \makecell{35.4\\{\scriptsize [33.4, 37.5]}} & \makecell{0.0129\\{\scriptsize [0.0044, 0.0214]}} & \makecell{0.756\\{\scriptsize [0.617, 0.894]}} & \makecell{93.3\\{\scriptsize [74.7, 112.0]}} \\
        DSGN-SCALAR & \makecell{0.611\\{\scriptsize [0.425, 0.797]}} & \makecell{34.0\\{\scriptsize [32.8, 35.2]}} & \makecell{0.0148\\{\scriptsize [0.0074, 0.0222]}} & \makecell{0.792\\{\scriptsize [0.777, 0.807]}} & \makecell{96.7\\{\scriptsize [79.3, 114.2]}} \\
        \makecell[l]{DSGN-SCALAR-\\WO-BETA} & \makecell{0.506\\{\scriptsize [0.241, 0.770]}} & \makecell{34.5\\{\scriptsize [33.7, 35.2]}} & \makecell{0.0223\\{\scriptsize [0.0095, 0.0351]}} & \makecell{0.810\\{\scriptsize [0.715, 0.905]}} & \makecell{125.8\\{\scriptsize [98.0, 153.7]}} \\
        \midrule
        \textbf{DSGN} & \makecell{\textbf{0.750}\\{\scriptsize [0.603, 0.897]}} & \makecell{35.1\\{\scriptsize [33.7, 36.6]}} & \makecell{\textbf{0.0080}\\{\scriptsize [0.0018, 0.0141]}} & \makecell{\textbf{0.751}\\{\scriptsize [0.657, 0.845]}} & \makecell{\textbf{85.8}\\{\scriptsize [82.6, 89.1]}} \\
        \bottomrule
    \end{tabular}}
    }
    \vspace{-3mm}
\end{table}

\subsubsection{Mode Switching Across Task Phases}

SSC requires reliable switching between surveillance and containment.
We evaluate this capability in two simulation settings: Pure Surveillance and Task Switching (Table~\ref{tab:ablation_transition}).
Pure Surveillance contains no evaders and requires the team to visit all surveillance nodes within the evaluation horizon.
Task Switching starts with the same surveillance objective, and evaders are later introduced near the boundary of the active pursuers' sensing range.
Success in this setting requires both complete surveillance-node coverage and successful evader containment, thereby testing the transition from global dispersion to tactical containment.

\begin{figure}[htbp]
    \centering
    \begin{tikzpicture}
        \node[anchor=south west,inner sep=0] (temporalplot) at (0,0)
            {\includegraphics[width=0.65\columnwidth]{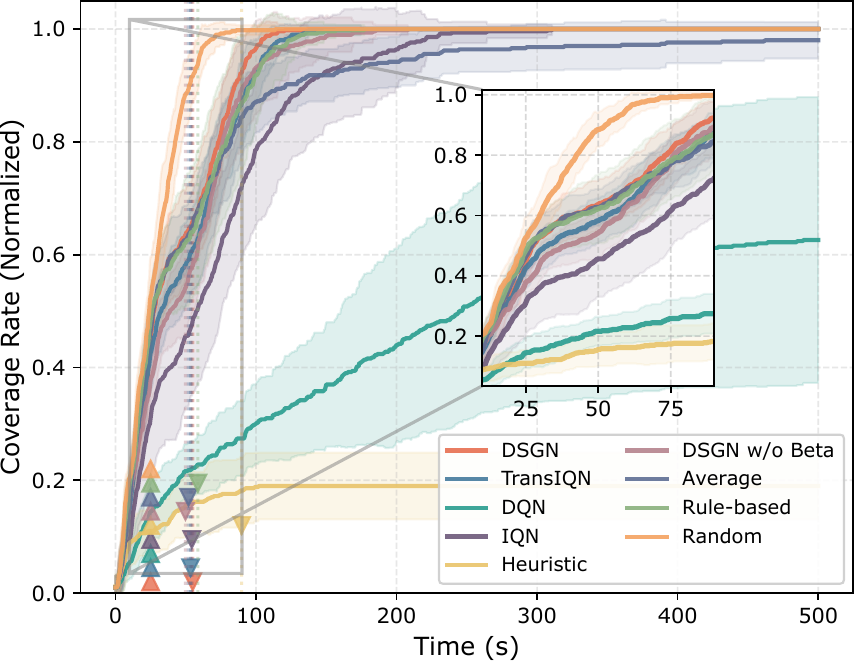}};
        \begin{scope}[x={(temporalplot.south east)},y={(temporalplot.north west)}]
            \fill[white] (0.77,0.298) rectangle (0.995,0.337);
            \node[anchor=west,inner sep=0,font=\sffamily\fontsize{4.1}{4.1}\selectfont]
                at (0.775,0.317) {DSGN-WO-BETA};
        \end{scope}
    \end{tikzpicture}
    \vspace{-2mm}
    \caption{Temporal coverage dynamics in the 6v2 task-switching scenario.}
    \label{fig:temporal_switching_6v2}
    \vspace{-2mm}
\end{figure}

\begin{table*}[htbp]
    \centering
    \color{black}
    \caption{\textcolor{black}{\textbf{Temporal Indicators for Mode Switching}}}
    \vspace{-1.5mm}
    \label{tab:temporal_switching_metrics}
    \footnotesize
    \setlength{\tabcolsep}{2.2pt}
    \resizebox{0.9\textwidth}{!}{%
    \renewcommand{\arraystretch}{1.0}
    \begin{tabular}{l|ccccc|ccccc}
        \toprule
        \multirow{2}{*}{\textbf{Scenario}}
        & \multicolumn{5}{c|}{\(\mathbf{C_{\text{switch}}}\) \(\uparrow\)}
        & \multicolumn{5}{c}{\(\mathbf{T_{\text{resume}}}\) (s) \(\downarrow\)} \\
        \cmidrule(lr){2-6} \cmidrule(lr){7-11}
        & {\scriptsize\textbf{DSGN}}
        & {\scriptsize\textbf{\shortstack{DSGN-WO-\\BETA}}}
        & {\scriptsize\textbf{\shortstack{DSGN-\\SCALAR}}}
        & {\scriptsize\textbf{\shortstack{DSGN-SCALAR-\\WO-BETA}}}
        & {\scriptsize\textbf{TransIQN}}
        & {\scriptsize\textbf{DSGN}}
        & {\scriptsize\textbf{\shortstack{DSGN-WO-\\BETA}}}
        & {\scriptsize\textbf{\shortstack{DSGN-\\SCALAR}}}
        & {\scriptsize\textbf{\shortstack{DSGN-SCALAR-\\WO-BETA}}}
        & {\scriptsize\textbf{TransIQN}} \\
        \midrule
        3v1
        & \makecell{\textbf{0.252}\\{\scriptsize \textbf{[0.216, 0.288]}}}
        & \makecell{0.232\\{\scriptsize [0.169, 0.294]}}
        & \makecell{0.234\\{\scriptsize [0.217, 0.251]}}
        & \makecell{0.240\\{\scriptsize [0.205, 0.275]}}
        & \makecell{0.237\\{\scriptsize [0.201, 0.273]}}
        & \makecell{\textbf{176.4}\\{\scriptsize \textbf{[83.1, 269.7]}}}
        & \makecell{212.4\\{\scriptsize [116.3, 308.5]}}
        & \makecell{250.5\\{\scriptsize [198.8, 302.2]}}
        & \makecell{269.5\\{\scriptsize [28.0, 511.1]}}
        & \makecell{268.5\\{\scriptsize [153.8, 383.1]}} \\
        6v2
        & \makecell{\textbf{0.466}\\{\scriptsize \textbf{[0.388, 0.544]}}}
        & \makecell{0.436\\{\scriptsize [0.329, 0.543]}}
        & \makecell{0.440\\{\scriptsize [0.405, 0.475]}}
        & \makecell{0.455\\{\scriptsize [0.395, 0.515]}}
        & \makecell{0.446\\{\scriptsize [0.409, 0.483]}}
        & \makecell{\textbf{55.5}\\{\scriptsize \textbf{[35.4, 75.7]}}}
        & \makecell{69.3\\{\scriptsize [39.5, 99.0]}}
        & \makecell{89.0\\{\scriptsize [65.3, 112.7]}}
        & \makecell{100.6\\{\scriptsize [53.8, 147.5]}}
        & \makecell{99.0\\{\scriptsize [26.9, 171.1]}} \\
        9v3
        & \makecell{\textbf{0.636}\\{\scriptsize \textbf{[0.551, 0.721]}}}
        & \makecell{0.581\\{\scriptsize [0.470, 0.691]}}
        & \makecell{0.610\\{\scriptsize [0.590, 0.630]}}
        & \makecell{0.620\\{\scriptsize [0.566, 0.673]}}
        & \makecell{0.608\\{\scriptsize [0.563, 0.654]}}
        & \makecell{\textbf{31.6}\\{\scriptsize \textbf{[20.9, 42.4]}}}
        & \makecell{36.3\\{\scriptsize [23.9, 48.8]}}
        & \makecell{41.4\\{\scriptsize [36.3, 46.5]}}
        & \makecell{45.9\\{\scriptsize [33.1, 58.8]}}
        & \makecell{56.9\\{\scriptsize [17.4, 96.5]}} \\
        \bottomrule
    \end{tabular}
    }

    \vspace{0.8mm}
    \begin{minipage}{\textwidth}
        \scriptsize
        \emph{Note:} \(C_{\text{switch}}\) denotes coverage at switching; \(T_{\text{resume}}\) denotes recovery time in seconds. Brackets denote 95\% confidence intervals. \textcolor{black}{\(T_{\text{resume}}\) is conditioned on episodes that complete both containment and final coverage.}
    \end{minipage}
    \vspace{-6mm}
\end{table*}

\textcolor{black}{Table~\ref{tab:ablation_transition} shows that DSGN has the shortest mean Task Switching episode duration in all settings and Pure Surveillance SR of 1.000 at every tested team size. Random's faster \(N=5\) patrol does not translate to reliable switching. Under scalar gating, the Beta prior improves Task Switching SR and shortens mean episode duration in every configuration, while raising Pure Surveillance SR at every tested team size.}

To further clarify the switching process, we analyze the temporal coverage curves using two quantities: the normalized coverage at the switching instant, denoted by \(C_{\text{switch}}\), and the post-containment surveillance recovery time, denoted by \(T_{\text{resume}}\), which measures the time from containment completion to final coverage completion.
\textcolor{black}{\(T_{\text{resume}}\) is computed only for episodes that complete both containment and final coverage; incomplete episodes are excluded rather than assigned the horizon value, while SR is computed over all episodes.}
Because coverage is cumulative, Fig.~\ref{fig:temporal_switching_6v2} reflects surveillance interruption and recovery rather than containment success alone.
The upward triangles mark evader spawning at \(t=25\) s, while the downward triangles and dotted lines mark average containment completion.
The coverage trajectory typically follows an ``increase--flatten--increase'' pattern: surveillance progress accumulates before the switching event, slows down during containment, and resumes after all evaders are contained.
Table~\ref{tab:temporal_switching_metrics} summarizes these temporal indicators.
\textcolor{black}{DSGN has the highest mean \(C_{\text{switch}}\) and shortest mean \(T_{\text{resume}}\) in all three settings (Table~\ref{tab:temporal_switching_metrics}). Its advantage is more pronounced in recovery time, indicating that DSGN chiefly improves how quickly surveillance resumes after containment.}

\subsubsection{Sensitivity to Search-Region Geometry}

\textcolor{black}{We evaluate 6v2 task switching under three search-region layouts: G0 has no holes, G2 has one central hexagonal hole, and G3 has four triangular holes (Fig.~\ref{fig:geometry_schematics} in Appendix~\ref{app:additional-results}). In both G2 and G3, the holes occupy 20.0\% of the workspace; the outer boundary is unchanged. Comparing these equal-area hole layouts tests sensitivity to spatial arrangement rather than blocked fraction.}
\textcolor{black}{DSGN has the highest mean SR and lowest mean P-Steps in each layout (Table~\ref{tab:geometry_ablation}), indicating the strongest combined task completion and post-containment surveillance despite not always leading in time or CR. Its SR still drops from 1.000 in G0 to 0.500 in G3, showing that complex layouts remain challenging despite its relative advantage.}

\subsubsection{Sensitivity to Action Granularity}\label{subsec:action_granularity}

\textcolor{black}{We compare separately trained DSGN policies with \(3\times3\) and \(5\times5\) command grids in 3v1, 6v2, and 9v3 Task Switching (Table~\ref{tab:action_granularity}). The finer grid adds intermediate commands within the same bounds.}
\textcolor{black}{The \(5\times5\) grid has mean \(T_{\mathrm{cap}}\) and P-Steps close to those of \(3\times3\), with slightly shorter containment times in 6v2 and 9v3 and fewer P-Steps in 3v1 and 6v2. Although its SR is lower and CR higher in all three settings, DSGN remains effective with the finer action grid, demonstrating good adaptability to higher-resolution action spaces while maintaining comparable execution efficiency.}

    \section{Conclusion}\label{sec:conclusion}
    DSGN separates surveillance and containment representations and coordinates them through vector-wise gating with a state-dependent Beta prior.
\textcolor{black}{Experiments show the benefits of DSGN for task completion and post-containment surveillance recovery at different team sizes.}
\textcolor{black}{Future work will investigate its application in continuous action spaces, repeated target-appearance cycles, and three-dimensional nonholonomic robots with limited communication.}

    \bibliographystyle{IEEEtranRAL}
    \bibliography{references}

    \onecolumn
    \appendices
    \section{Additional Experimental Results}\label{app:additional-results}
    \setcounter{table}{0}
    \setcounter{figure}{0}
    \renewcommand{\thetable}{A.\Roman{table}}
    \renewcommand{\thefigure}{A.\arabic{figure}}
    \renewcommand{\theHtable}{appendix.A.\arabic{table}}
    \renewcommand{\theHfigure}{appendix.A.\arabic{figure}}
\begin{table}[H]
\centering
\color{black}
\caption{\textbf{Comparison of Arbitration Variants across Small, Medium, and Large Scales}}
\label{tab:ablation_gating}
\scriptsize
\setlength{\tabcolsep}{3pt}
\renewcommand{\arraystretch}{1.12}
\par\smallskip\textbf{(a) Small scale: 12v3--20v5}\par\smallskip
\begin{adjustbox}{width=0.75\linewidth,center}
\begin{tabular}{@{}lccccc@{}}
\toprule
\textbf{Variant} & SR $\uparrow$ & $T_{\mathrm{cap}}$ (s) $\downarrow$ & CR $\downarrow$ & Var $\downarrow$ & P-Steps $\downarrow$ \\
\midrule
Random & \makecell{0.433\\{\scriptsize [0.253, 0.614]}} & \makecell{128.4\\{\scriptsize [83.4, 173.4]}} & \makecell{0.0543\\{\scriptsize [0.0317, 0.0770]}} & \makecell{2.267\\{\scriptsize [2.050, 2.484]}} & \makecell{87.9\\{\scriptsize [77.0, 98.8]}} \\
Average & \makecell{0.917\\{\scriptsize [0.846, 0.987]}} & \makecell{\textbf{23.5}\\{\scriptsize [21.8, 25.1]}} & \makecell{0.0104\\{\scriptsize [0.0012, 0.0197]}} & \makecell{0.839\\{\scriptsize [0.707, 0.970]}} & \makecell{84.8\\{\scriptsize [66.4, 103.1]}} \\
Rule-based & \makecell{0.800\\{\scriptsize [0.698, 0.902]}} & \makecell{32.4\\{\scriptsize [30.2, 34.6]}} & \makecell{0.0245\\{\scriptsize [0.0108, 0.0383]}} & \makecell{1.290\\{\scriptsize [1.155, 1.425]}} & \makecell{88.0\\{\scriptsize [81.4, 94.7]}} \\
\midrule
DSGN-WO-BETA & \makecell{0.878\\{\scriptsize [0.771, 0.985]}} & \makecell{24.0\\{\scriptsize [22.8, 25.3]}} & \makecell{0.0137\\{\scriptsize [0.0016, 0.0258]}} & \makecell{0.848\\{\scriptsize [0.786, 0.910]}} & \makecell{78.5\\{\scriptsize [74.7, 82.4]}} \\
DSGN-SCALAR & \makecell{0.911\\{\scriptsize [0.826, 0.996]}} & \makecell{23.7\\{\scriptsize [22.8, 24.6]}} & \makecell{0.0089\\{\scriptsize [0.0011, 0.0168]}} & \makecell{0.870\\{\scriptsize [0.777, 0.963]}} & \makecell{75.3\\{\scriptsize [67.2, 83.3]}} \\
\makecell[l]{DSGN-SCALAR-\\WO-BETA} & \makecell{0.850\\{\scriptsize [0.726, 0.974]}} & \makecell{24.1\\{\scriptsize [23.8, 24.4]}} & \makecell{0.0168\\{\scriptsize [0.0071, 0.0265]}} & \makecell{0.830\\{\scriptsize [0.786, 0.875]}} & \makecell{90.4\\{\scriptsize [77.4, 103.3]}} \\
\midrule
\textbf{DSGN} & \makecell{\textbf{0.922}\\{\scriptsize [0.865, 0.980]}} & \makecell{23.6\\{\scriptsize [23.3, 23.9]}} & \makecell{\textbf{0.0083}\\{\scriptsize [0.0007, 0.0159]}} & \makecell{\textbf{0.789}\\{\scriptsize [0.722, 0.856]}} & \makecell{\textbf{70.7}\\{\scriptsize [64.6, 76.7]}} \\
\bottomrule
\end{tabular}
\end{adjustbox}
\par\smallskip\textbf{(b) Medium scale: 38v10--64v16}\par\smallskip
\begin{adjustbox}{width=0.75\linewidth,center}
\begin{tabular}{@{}lccccc@{}}
\toprule
\textbf{Variant} & SR $\uparrow$ & $T_{\mathrm{cap}}$ (s) $\downarrow$ & CR $\downarrow$ & Var $\downarrow$ & P-Steps $\downarrow$ \\
\midrule
Random & \makecell{0.000\\{\scriptsize [0.000, 0.000]}} & \makecell{356.4\\{\scriptsize [261.5, 451.3]}} & \makecell{0.1426\\{\scriptsize [0.1142, 0.1709]}} & \makecell{2.442\\{\scriptsize [2.299, 2.585]}} & \makecell{118.7\\{\scriptsize [103.8, 133.7]}} \\
Average & \makecell{0.733\\{\scriptsize [0.620, 0.846]}} & \makecell{\textbf{32.3}\\{\scriptsize [29.9, 34.7]}} & \makecell{0.0085\\{\scriptsize [0.0043, 0.0128]}} & \makecell{0.770\\{\scriptsize [0.683, 0.858]}} & \makecell{82.0\\{\scriptsize [76.8, 87.2]}} \\
Rule-based & \makecell{0.517\\{\scriptsize [0.389, 0.644]}} & \makecell{44.5\\{\scriptsize [39.8, 49.2]}} & \makecell{0.0298\\{\scriptsize [0.0140, 0.0455]}} & \makecell{1.042\\{\scriptsize [0.944, 1.140]}} & \makecell{105.4\\{\scriptsize [96.3, 114.6]}} \\
\midrule
DSGN-WO-BETA & \makecell{0.711\\{\scriptsize [0.607, 0.815]}} & \makecell{33.5\\{\scriptsize [31.9, 35.0]}} & \makecell{0.0118\\{\scriptsize [0.0055, 0.0181]}} & \makecell{0.741\\{\scriptsize [0.624, 0.859]}} & \makecell{83.7\\{\scriptsize [72.8, 94.5]}} \\
DSGN-SCALAR & \makecell{0.750\\{\scriptsize [0.668, 0.832]}} & \makecell{32.9\\{\scriptsize [31.7, 34.2]}} & \makecell{0.0124\\{\scriptsize [0.0101, 0.0146]}} & \makecell{0.765\\{\scriptsize [0.652, 0.878]}} & \makecell{85.0\\{\scriptsize [74.3, 95.8]}} \\
\makecell[l]{DSGN-SCALAR-\\WO-BETA} & \makecell{0.606\\{\scriptsize [0.397, 0.814]}} & \makecell{32.4\\{\scriptsize [32.1, 32.7]}} & \makecell{0.0155\\{\scriptsize [0.0072, 0.0238]}} & \makecell{0.773\\{\scriptsize [0.728, 0.818]}} & \makecell{108.3\\{\scriptsize [88.5, 128.1]}} \\
\midrule
\textbf{DSGN} & \makecell{\textbf{0.811}\\{\scriptsize [0.637, 0.985]}} & \makecell{33.1\\{\scriptsize [31.8, 34.4]}} & \makecell{\textbf{0.0062}\\{\scriptsize [0.0016, 0.0108]}} & \makecell{\textbf{0.727}\\{\scriptsize [0.710, 0.744]}} & \makecell{\textbf{80.5}\\{\scriptsize [79.1, 81.8]}} \\
\bottomrule
\end{tabular}
\end{adjustbox}
\par\smallskip\textbf{(c) Large scale: 46v12--77v19}\par\smallskip
\begin{adjustbox}{width=0.75\linewidth,center}
\begin{tabular}{@{}lccccc@{}}
\toprule
\textbf{Variant} & SR $\uparrow$ & $T_{\mathrm{cap}}$ (s) $\downarrow$ & CR $\downarrow$ & Var $\downarrow$ & P-Steps $\downarrow$ \\
\midrule
Random & \makecell{0.000\\{\scriptsize [0.000, 0.000]}} & \makecell{380.9\\{\scriptsize [303.6, 458.1]}} & \makecell{0.1403\\{\scriptsize [0.1173, 0.1634]}} & \makecell{2.392\\{\scriptsize [2.276, 2.507]}} & \makecell{124.8\\{\scriptsize [113.8, 135.8]}} \\
Average & \makecell{\textbf{0.750}\\{\scriptsize [0.640, 0.860]}} & \makecell{\textbf{33.0}\\{\scriptsize [30.8, 35.2]}} & \makecell{0.0083\\{\scriptsize [0.0044, 0.0123]}} & \makecell{0.836\\{\scriptsize [0.752, 0.919]}} & \makecell{109.4\\{\scriptsize [85.9, 133.0]}} \\
Rule-based & \makecell{0.400\\{\scriptsize [0.275, 0.525]}} & \makecell{47.6\\{\scriptsize [43.5, 51.6]}} & \makecell{0.0320\\{\scriptsize [0.0219, 0.0422]}} & \makecell{1.056\\{\scriptsize [0.972, 1.140]}} & \makecell{113.6\\{\scriptsize [105.2, 122.0]}} \\
\midrule
DSGN-WO-BETA & \makecell{0.656\\{\scriptsize [0.540, 0.771]}} & \makecell{35.4\\{\scriptsize [33.4, 37.5]}} & \makecell{0.0129\\{\scriptsize [0.0044, 0.0214]}} & \makecell{0.756\\{\scriptsize [0.617, 0.894]}} & \makecell{93.3\\{\scriptsize [74.7, 112.0]}} \\
DSGN-SCALAR & \makecell{0.611\\{\scriptsize [0.425, 0.797]}} & \makecell{34.0\\{\scriptsize [32.8, 35.2]}} & \makecell{0.0148\\{\scriptsize [0.0074, 0.0222]}} & \makecell{0.792\\{\scriptsize [0.777, 0.807]}} & \makecell{96.7\\{\scriptsize [79.3, 114.2]}} \\
\makecell[l]{DSGN-SCALAR-\\WO-BETA} & \makecell{0.506\\{\scriptsize [0.241, 0.770]}} & \makecell{34.5\\{\scriptsize [33.7, 35.2]}} & \makecell{0.0223\\{\scriptsize [0.0095, 0.0351]}} & \makecell{0.810\\{\scriptsize [0.715, 0.905]}} & \makecell{125.8\\{\scriptsize [98.0, 153.7]}} \\
\midrule
\textbf{DSGN} & \makecell{\textbf{0.750}\\{\scriptsize [0.603, 0.897]}} & \makecell{35.1\\{\scriptsize [33.7, 36.6]}} & \makecell{\textbf{0.0080}\\{\scriptsize [0.0018, 0.0141]}} & \makecell{\textbf{0.751}\\{\scriptsize [0.657, 0.845]}} & \makecell{\textbf{85.8}\\{\scriptsize [82.6, 89.1]}} \\
\bottomrule
\end{tabular}
\end{adjustbox}
\par\smallskip
\raggedright\scriptsize\emph{Note:} Brackets denote 95\% confidence intervals. $T_{\mathrm{cap}}$ is conditioned on containment completion; P-Steps is conditioned on successful containment and final coverage.
\end{table}

\twocolumn

\begin{figure}[H]
    \centering
    \includegraphics[width=0.44\columnwidth]{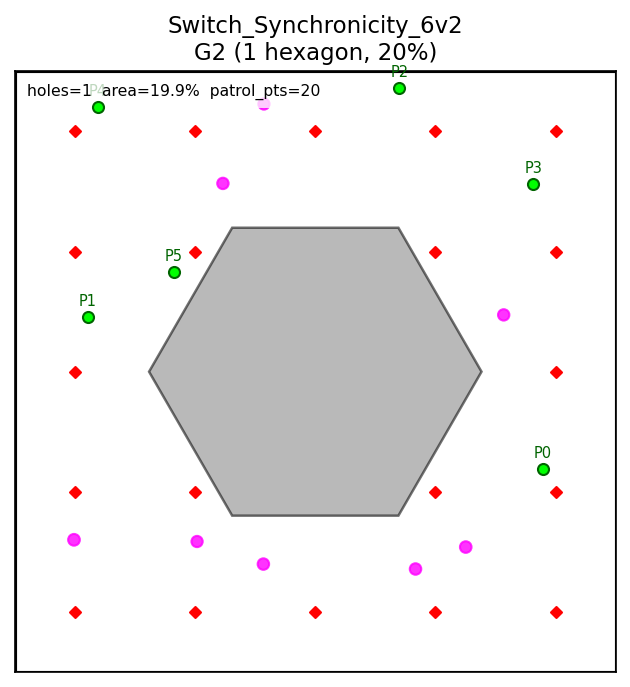}\hfill
    \includegraphics[width=0.44\columnwidth]{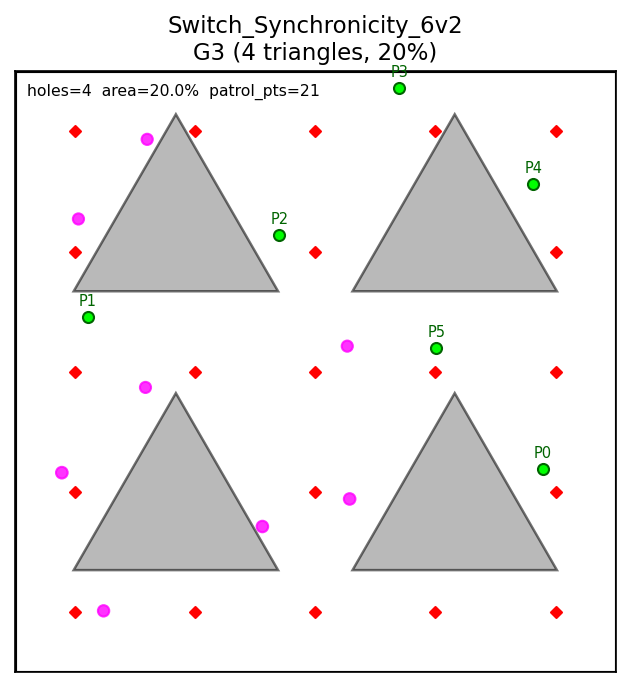}
    \caption{\textcolor{black}{Representative 6v2 layouts: G2 (one hexagonal hole, left) and G3 (four triangular holes, right). Green circles labeled P0--P5 are pursuers, magenta circles are obstacles, and red diamonds are surveillance nodes. Gray areas are inaccessible. G0 has no hole.}}
    \label{fig:geometry_schematics}

\end{figure}

\begin{table}[H]
    \centering
    \color{black}
    \caption{\textbf{6v2 Geometry Sensitivity across DSGN Variants and TransIQN}}
    \label{tab:geometry_ablation}
    \scriptsize
    \setlength{\tabcolsep}{1.5pt}
    \renewcommand{\arraystretch}{1.0}
    \newcommand{\geoCI}[3]{\makecell{#1\\{\scriptsize [#2,#3]}}}
    \textit{(a) G0: no hole}
    \vspace{0.5mm}

    \resizebox{0.95\columnwidth}{!}{%
    \begin{tabular}{@{}lcccc@{}}
        \toprule
        \textbf{Method} & \textbf{SR}$\uparrow$ & $\mathbf{T_{\mathrm{cap}}}$ (s)$\downarrow$ & \textbf{CR}$\downarrow$ & \textbf{P-Steps}$\downarrow$ \\
        \midrule
        TransIQN & \geoCI{0.950}{0.765}{0.990} & \geoCI{\textbf{26.1}}{20.8}{31.4} & \geoCI{0.0083}{0.0000}{0.0258} & \geoCI{227.8}{168.9}{286.7} \\
        \midrule
        D-w/o-B & \geoCI{0.935}{0.790}{1.000} & \geoCI{30.3}{25.0}{35.6} & \geoCI{0.0083}{0.0000}{0.0290} & \geoCI{129.6}{109.4}{149.8} \\
        D-S & \geoCI{0.965}{0.825}{1.000} & \geoCI{28.7}{20.6}{36.8} & \geoCI{0.0111}{0.0000}{0.0589} & \geoCI{186.3}{86.2}{286.3} \\
        D-S-w/o-B & \geoCI{0.885}{0.485}{1.000} & \geoCI{30.0}{23.2}{36.8} & \geoCI{0.0083}{0.0000}{0.0290} & \geoCI{229.3}{186.8}{271.7} \\
        \midrule
        \textbf{DSGN} & \geoCI{\textbf{1.000}}{1.000}{1.000} & \geoCI{27.1}{18.4}{35.8} & \geoCI{\textbf{0.0000}}{0.0000}{0.0000} & \geoCI{\textbf{89.7}}{40.5}{139.0} \\
        \bottomrule
    \end{tabular}
    }

    \vspace{1.2mm}
    \textit{(b) G2: one hexagonal hole}
    \vspace{0.5mm}

    \resizebox{0.95\columnwidth}{!}{%
    \begin{tabular}{@{}lcccc@{}}
        \toprule
        \textbf{Method} & \textbf{SR}$\uparrow$ & $\mathbf{T_{\mathrm{cap}}}$ (s)$\downarrow$ & \textbf{CR}$\downarrow$ & \textbf{P-Steps}$\downarrow$ \\
        \midrule
        TransIQN & \geoCI{0.650}{0.435}{0.820} & \geoCI{\textbf{42.0}}{29.3}{54.7} & \geoCI{0.0833}{0.0188}{0.1479} & \geoCI{244.6}{163.9}{325.3} \\
        \midrule
        D-w/o-B & \geoCI{0.835}{0.435}{1.000} & \geoCI{55.0}{52.6}{57.4} & \geoCI{0.0444}{0.0000}{0.1640} & \geoCI{107.5}{85.1}{129.8} \\
        D-S & \geoCI{0.815}{0.560}{1.000} & \geoCI{44.6}{36.0}{53.1} & \geoCI{0.0583}{0.0000}{0.1532} & \geoCI{144.8}{137.1}{152.4} \\
        D-S-w/o-B & \geoCI{0.765}{0.510}{1.000} & \geoCI{52.4}{10.2}{94.6} & \geoCI{0.0556}{0.0239}{0.0872} & \geoCI{177.9}{146.4}{209.4} \\
        \midrule
        \textbf{DSGN} & \geoCI{\textbf{0.865}}{0.675}{1.000} & \geoCI{43.3}{27.4}{59.2} & \geoCI{\textbf{0.0361}}{0.0242}{0.0481} & \geoCI{\textbf{78.5}}{25.2}{131.8} \\
        \bottomrule
        \end{tabular}
    }

    \vspace{1.2mm}
    \textit{(c) G3: four triangular holes}
    \vspace{0.5mm}

    \resizebox{0.95\columnwidth}{!}{%
    \begin{tabular}{@{}lcccc@{}}
        \toprule
        \textbf{Method} & \textbf{SR}$\uparrow$ & $\mathbf{T_{\mathrm{cap}}}$ (s)$\downarrow$ & \textbf{CR}$\downarrow$ & \textbf{P-Steps}$\downarrow$ \\
        \midrule
        TransIQN & \geoCI{0.350}{0.180}{0.565} & \geoCI{\textbf{61.5}}{34.8}{88.2} & \geoCI{0.1583}{0.0689}{0.2477} & \geoCI{388.0}{299.6}{476.4} \\
        \midrule
        D-w/o-B & \geoCI{0.485}{0.105}{0.865} & \geoCI{93.1}{43.3}{142.8} & \geoCI{0.1361}{0.0696}{0.2027} & \geoCI{154.5}{78.7}{230.3} \\
        D-S & \geoCI{0.350}{0.230}{0.470} & \geoCI{65.7}{33.8}{97.6} & \geoCI{0.1444}{0.0000}{0.2942} & \geoCI{236.6}{128.9}{344.2} \\
        D-S-w/o-B & \geoCI{0.450}{0.260}{0.660} & \geoCI{68.7}{0.0}{160.1} & \geoCI{\textbf{0.1139}}{0.0506}{0.1771} & \geoCI{363.8}{263.5}{464.2} \\
        \midrule
        \textbf{DSGN} & \geoCI{\textbf{0.500}}{0.375}{0.625} & \geoCI{65.6}{36.7}{94.5} & \geoCI{0.1278}{0.0000}{0.3201} & \geoCI{\textbf{148.8}}{81.6}{215.9} \\
        \bottomrule
    \end{tabular}
    }
    \par\vspace{1mm}
    \noindent\begin{minipage}{\columnwidth}
        \scriptsize
        \emph{Note:} D-w/o-B, D-S, and D-S-w/o-B denote DSGN-WO-BETA, DSGN-SCALAR, and DSGN-SCALAR-WO-BETA, respectively. Brackets show 95\% confidence intervals.
    \end{minipage}

\end{table}

\newpage

\begin{table}[H]
    \centering
    \color{black}
    \caption{\textbf{DSGN Action-Granularity Sensitivity in Task Switching}}
    \label{tab:action_granularity}
    \footnotesize
    \setlength{\tabcolsep}{1.5pt}
    \renewcommand{\arraystretch}{1.05}
    \newcommand{\gridCI}[3]{\makecell{#1\\{\scriptsize [#2,#3]}}}
    \resizebox{0.95\columnwidth}{!}{%
    \begin{tabular}{@{}llcccc@{}}
        \toprule
        \textbf{Scenario} & \textbf{Grid} & \textbf{SR}$\uparrow$ & $\mathbf{T_{\mathrm{cap}}}$ (s)$\downarrow$ & \textbf{CR}$\downarrow$ & \textbf{P-Steps}$\downarrow$ \\
        \midrule
        \multirow{2}{*}{3v1} & \(3\times3\) & \gridCI{\textbf{0.995}}{0.985}{1.000} &  \gridCI{\textbf{24.2}}{21.8}{26.6}& \gridCI{\textbf{0.0000}}{0.0000}{0.0000} & \gridCI{218.2}{210.0}{226.4} \\
                             & \(5\times5\) & \gridCI{0.980}{0.961}{0.999} & \gridCI{24.5}{21.9}{27.1} & \gridCI{0.0083}{0.0000}{0.0169} & \gridCI{\textbf{216.0}}{209.1}{222.9} \\
        \multirow{2}{*}{6v2} & \(3\times3\) & \gridCI{\textbf{0.970}}{0.946}{0.994} & \gridCI{33.7}{31.7}{35.7} & \gridCI{\textbf{0.0075}}{0.0012}{0.0138} &   \gridCI{88.8}{84.5}{93.1} \\
                             & \(5\times5\) & \gridCI{0.955}{0.926}{0.984} & \gridCI{\textbf{32.9}}{31.2}{34.6} & \gridCI{0.0108}{0.0034}{0.0182} & \gridCI{\textbf{86.2}}{82.2}{90.2} \\
        \multirow{2}{*}{9v3} & \(3\times3\) & \gridCI{\textbf{0.900}}{0.858}{0.942} & \gridCI{28.8}{27.2}{30.4} & \gridCI{\textbf{0.0128}}{0.0072}{0.0184} &  \gridCI{\textbf{52.1}}{48.6}{55.6} \\
                             & \(5\times5\) & \gridCI{0.880}{0.835}{0.925} & \gridCI{\textbf{27.6}}{26.0}{29.2} & \gridCI{0.0161}{0.0097}{0.0225} &  \gridCI{52.9}{49.4}{56.4} \\
        \bottomrule
    \end{tabular}}
    \par\vspace{1mm}
    \noindent\begin{minipage}{\columnwidth}
        \scriptsize
        \emph{Note:} Mean with 95\% confidence interval in brackets; bold denotes the best mean within each scenario (ties included). The \(5\times5\) grid uses \(a_{\mathrm{acc}}\in\{0,\pm0.2,\pm0.4\}\,\mathrm{m/s}^2\) and \(\omega\in\{0,\pm\pi/12,\pm\pi/6\}\,\mathrm{rad/s}\). 
    \end{minipage}

\end{table}

\end{document}